\documentclass[acmsmall]{acmart}
\AtBeginDocument{%
  }

\setcopyright{cc}
\setcctype{by}
\acmJournal{PACMMOD}
\acmYear{2025} \acmVolume{3} \acmNumber{3 (SIGMOD)} \acmArticle{209} \acmMonth{6} \acmPrice{}\acmDOI{10.1145/3725346}

\newcommand\modelName{RLOMM}
\usepackage{tikz}
\usepackage{enumitem}
\usepackage[ruled,vlined,linesnumbered]{algorithm2e}
\usepackage{algpseudocode}
\usepackage{multirow}
\usepackage{multicol}
\usepackage{stfloats} 
\usepackage{booktabs}
\usepackage{graphicx} 

\usepackage{array}
\usepackage{svg}
\usepackage{subcaption}
\usepackage{pgfplots}
\usepackage{verbatim}
\usepackage{amsmath,bm}

\usepackage{tkz-euclide}
\usepackage{marginnote}
\usepackage{tcolorbox}
\tcbuselibrary{breakable}
\usepackage{diagbox}
\usepackage{xcolor}

\begin{document}

\title{RLOMM: An Efficient and Robust Online Map Matching Framework with Reinforcement Learning}

\author{Minxiao Chen}
\authornote{Both authors contributed equally to this research.}
\affiliation{%
  \institution{Beijing University of Posts and Telecommunications}
  \country{China}
}
\affiliation{%
  \institution{Beiyou Shenzhen Institute}
  \country{China}
}
\email{chenminxiao@bupt.edu.cn}

\author{Haitao Yuan}
\authornotemark[1]
\affiliation{%
  \institution{Nanyang Technological University}
  \country{Singapore}
}
\email{haitao.yuan@ntu.edu.sg}

\author{Nan Jiang}
\affiliation{%
  \institution{Beijing University of Posts and Telecommunications}
  \country{China}
}
\email{jn_bupt@bupt.edu.cn}

\author{Zhihan Zheng}
\affiliation{%
  \institution{Beijing University of Posts and Telecommunications}
  \country{China}
}
\email{26510036@bupt.edu.cn}

\author{Sai Wu}
\affiliation{%
  \institution{Zhejiang University}
  \country{China}
}
\email{wusai@zju.edu.cn}

\author{Ao Zhou}
\affiliation{%
  \institution{Beijing University of Posts and Telecommunications}
  \country{China}
}
\email{aozhou@bupt.edu.cn}

\author{Shangguang Wang}
\affiliation{%
  \institution{Beiyou Shenzhen Institute}
  \country{China}
}
\affiliation{%
  \institution{Beijing University of Posts and Telecommunications}
  \country{China}
}
\email{sgwang@bupt.edu.cn}

\renewcommand{\shortauthors}{Minxiao Chen, Haitao Yuan, Nan Jiang, Zhihan Zheng, Sai Wu, Ao Zhou and Shangguang Wang}

\begin{abstract}
Online map matching is a fundamental problem in location-based services, aiming to incrementally match trajectory data step-by-step onto a road network. However, existing methods fail to meet the needs for efficiency, robustness, and accuracy required by large-scale online applications, making this task still challenging. This paper introduces a novel framework that achieves high accuracy and efficient matching while ensuring robustness in handling diverse scenarios. To improve efficiency, we begin by modeling the online map matching problem as an Online Markov Decision Process (OMDP) based on its inherent characteristics. This approach helps efficiently merge historical and real-time data, reducing unnecessary calculations. Next, to enhance robustness, we design a reinforcement learning method, enabling robust handling of real-time data from dynamically changing environments. In particular, we propose a novel model learning process and a comprehensive reward function, allowing the model to make reasonable current matches from a future-oriented perspective, and to continuously update and optimize during the decision-making process based on feedback. Lastly, to address the heterogeneity between trajectories and roads, we design distinct graph structures, facilitating efficient representation learning through graph and recurrent neural networks. To further align trajectory and road data, we introduce contrastive learning to decrease their distance in the latent space, thereby promoting effective integration of the two. Extensive evaluations on three real-world datasets confirm that our method significantly outperforms existing state-of-the-art solutions in terms of accuracy, efficiency and robustness.
\end{abstract}




\received{October 2024}
\received[revised]{January 2025}
\received[accepted]{February 2025}

\maketitle

\section{Introduction}
Since the widespread availability of GPS services in the 1990s, research related to map matching has been a focal point. Map matching is the process of aligning vehicle trajectories onto the actual road network, which is a fundamental problem in location-based services~\cite{survey_magzine, stsurvey}. 
Online map matching is a sequential process that aligns incoming data incrementally, making it particularly well-suited for real-time applications. This process is a crucial element for various location-based online services and is vital for applications requiring real-time responses such as navigation services~\cite{navigation2}, optimal route planning~\cite{Learning2Route, AnySto}, and travel time estimation~\cite{TTE, TTE2, TTE3}.

The map matching method can be categorized into two main types: \textit{rule-based}~\cite{HMM,FMM,MDPMM} and \textit{deep-learning-based}~\cite{deepmmzhao, deepmmfeng, mtrajrec, l2mm, graphmm}. The former treats trajectories as observations and utilizes the Hidden Markov Model (HMM) to infer road segments as hidden states~\cite{HMMbase} or uses Markov Decision Process (MDP) with dynamic programming algorithms like value iteration to model offline matching process~\cite{MDPMM}. With advancements in deep learning, attention has shifted towards the latter approach, framing the problem as an end-to-end task. This method leverages large datasets to directly learn patterns for matching trajectories to road segments. Overall, both types of methods perform well in terms of matching accuracy in high-sampling-rate offline scenarios. However, when the problem shifts to the online scenario, they exhibit significant inefficiencies due to their inherent nature~\cite{praGuide, hmmAna, mmsurvey,surveyVMMT}.

Specifically, for online map matching scenarios, existing solutions are essentially straightforward extended forms of offline methods. For example, \cite{HMM} points out that Microsoft treats the map matching problem as a batch processing task, performing matching after all data has been collected. For online scenarios, they simply use a sliding window as an adaptation of the offline method. To offer a clear perspective, we summarize the two existing extension strategies, as shown in Fig.~\ref{fig:intro}. The first strategy involves repeatedly invoking the DNN or MDP-based offline methods~\cite{mtrajrec,l2mm, graphmm,MDPMM} during online matching process to match the current available trajectory at that moment. The second strategy employs HMM, incrementally matching incoming trajectories in a rule-based manner~\cite{onlinelearning, AMM}. However, current service providers still incur substantial costs because these two categories have the following notable deficiencies in addressing the core challenges of online map matching:

\begin{figure}
  \centering  
  \includegraphics[width=0.7\linewidth]{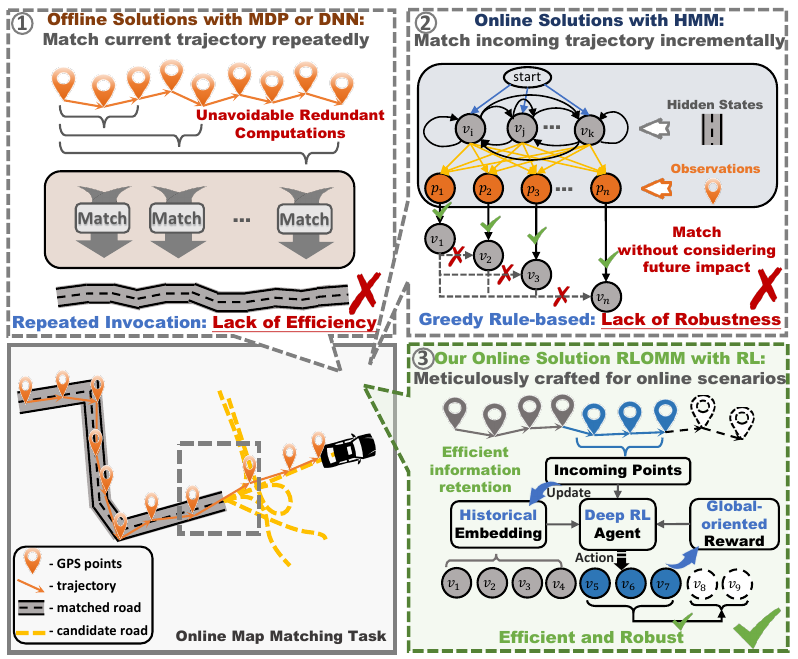}
  \vspace{-0.1in}
  \caption{Concept comparison of existing online map matching solutions and our \modelName.}
  \vspace{-0.25in}
  \label{fig:intro}
\end{figure}

\noindent (1) \textbf{Low efficiency.} Most current methods, particularly \textcolor{black}{DNN-based or MDP-based}, are tailored to align complete trajectories with road networks. However, when applied incrementally, these methods necessitate the input of all preceding trajectory segments to utilize historical data for enhancing effectiveness. This process results in redundant computations and leads to significant inefficiencies. In light of this issue, it is crucial to develop an innovative solution that not only maintains the accuracy of map matching but also enhances efficiency in real-time scenarios, thereby reducing costs for service providers.

\noindent (2) \textbf{Poor robustness.} Online map matching necessitates the continuous processing of data within a streaming pipeline, emphasizing the significant influence of previous matches on subsequent ones. Earlier matches create an informational context that shapes later decisions. However, existing online methods predominantly utilize a greedy approach, neglecting the essential impact of current matches on future results, thus leading to suboptimal matches. Consequently, addressing this oversight is crucial for advancing the development of a robust online map matching framework.

\noindent (3) \textbf{Insufficient handling of trajectory-road heterogeneity.} \textcolor{black}{Intuitively, trajectory data is an unstructured sequence with significant noise, whereas road network data is a relatively fixed, structured topological graph.} This means trajectory and road data are inherently heterogeneous, which demands careful address. Correlations exist both among different trajectories and between various road segments, all of which should be considered. However, existing methods lack specialized designs that effectively capture these complex correlations within and between data types, leading to suboptimal data representations and encoding techniques. We argue that an ideal solution should facilitate the integration of trajectory and road data while preserving their unique attributes and contexts. 

To comprehensively address the aforementioned limitations of existing methods, \textcolor{black}{as shown in Fig.~\ref{fig:intro}}, we introduce \textbf{\modelName}, an efficient and robust online map matching framework with high dynamic adaptability, which is a novel paradigm to address the online map matching problem.
\textbf{First, we meticulously design an Online Markov Decision Process (OMDP) specifically for online map matching scenario,} continuously extracting real-time and historical information from real-time environment to construct the information state, closely integrating with the incremental and streaming characteristics of online map matching. 
Compared to the MDP method previously used for offline map matching~\cite{MDPMM}, this novel modeling approach effectively captures the essential data and dynamically updates the historical information throughout the matching process, which fundamentally simplifies the complexity of the problem and prevents redundant calculations, thereby enhancing efficiency. Additionally, the efficient component that captures sequence correlations effectively coordinates with this modeling approach, enhancing the integration efficiency of historical and real-time information.

\textbf{Second, to endow the model with sufficient dynamic adaptability and robustness for various scenarios,} we employ reinforcement learning to execute matching actions based on the states divided by the OMDP and meticulously design the reward evaluation process. Specifically, we employ the deep Q-learning method, which allows the model to consider the impact of each match on future matches. This enables the model to perform matching from a future-oriented perspective, avoiding current optimal matches that are prone to errors, thereby discarding the simplistic greedy approach to achieve ideal overall matching accuracy. Meanwhile, a carefully designed training process enables the model to grasp the dynamic changes of the matching environment and allows for continuous updates and improvements in the decision-making process based on feedback from the rewards.

\textbf{Third, considering the trajectory-road heterogeneity,} we design distinct graph data structures for each type of data to capture their intrinsic connections, namely trajectory transition graph and link connection graph. By representing both types of data in effective graph structures, their heterogeneity can be mitigated, thus promoting better information integration. Additionally, we design mechanism to bridge the interaction between their representation to facilitate better fusion. Moreover, corresponding graph and recurrent neural networks are utilized to generate effective representations for both trajectories and roads. Recognizing the necessity for effective alignment and integration of trajectory data with road data in map matching task, we design a trajectory-road representation alignment module from the perspective of enhancing representational robustness. This module reduces the distance between the representations of the two data types in the latent space, thereby facilitating effective integration.

In summary, we make the following contributions:
\vspace{-\topsep}
\begin{itemize}[leftmargin=10.2pt]
\setlength{\itemsep}{0pt}
\setlength{\parsep}{0pt}
\setlength{\parskip}{0pt}
\item To the best of our knowledge, we are the first to model the online map matching problem as an Online Markov Decision Process, proposing a new solution paradigm. Additionally, we propose \textbf{\modelName}, a novel two-stage system framework based on deep reinforcement learning, which delivers efficient map matching with high dynamic adaptability and robustness. (Sec.~\ref{sec:2}\&\ref{sec:3})
\item We design graph structures meticulously tailored for both trajectory and road to capture the correlations inherent to each type of data and mitigate the heterogeneity. Our feature encoding modules utilize various representation learning and sequence correlation capture techniques to integrate trajectory and road data, facilitating the map matching process. (Sec.~\ref{sec:4})
\item We propose a novel model learning process for our \textbf{\modelName} framework with carefully designed reward evaluation, which can perform globally optimal matching from a future-oriented perspective. We introduce a trajectory-road representation alignment module that utilizes contrastive learning to facilitate the effective integration of trajectories and roads. Combined with the temporal-difference loss of reinforcement, our framework is capable of robust learning from a variety of scenarios. (Sec.~\ref{sec:5})
\item We conduct a comprehensive evaluation on three real-world datasets. The results show that our method significantly outperforms state-of-the-art in terms of accuracy, efficiency and robustness. (Sec.~\ref{sec:6})
\end{itemize}

\section{Preliminaries}
\label{sec:2}
\begin{figure}
  \centering  
  \includegraphics[width=0.7\linewidth]{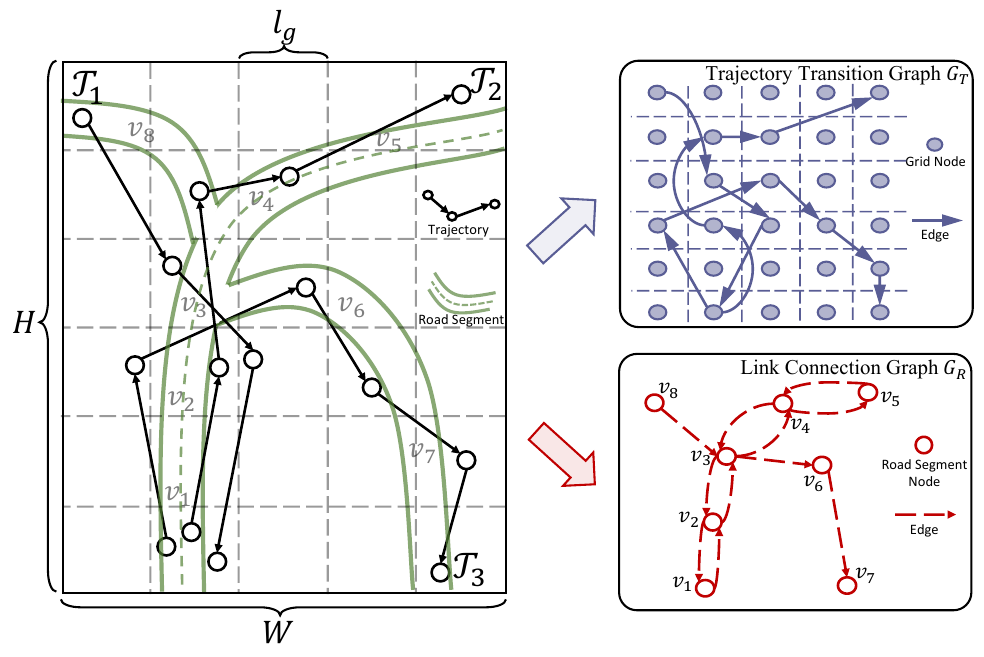}
  \vspace{-0.15in}
  \caption{An illustration of Grids Partition, Trajectory Transition Graph, and Link Connection Graph.}
  \vspace{-0.25in}
  \label{fig:pre}
\end{figure}


We first introduce the basic concepts of the map matching problem, namely trajectories and road networks, followed by a detailed explanation of the specifically designed data structures for this task. Finally, we formalize the online map matching problem.

\subsection{Basic Concepts}

\noindent \textbf{Trajectory.} A trajectory $\mathcal{T}$ is a sequence of GPS points, denoted as $\mathcal{T} = \langle p_1, p_2, \ldots, p_n \rangle$, where each point $p_i$ represents a location record captured at a specific time. Each point $p_i$ consists of three components: the geographic coordinates latitude ($\text{lat}_i$) and longitude ($\text{lon}_i$), and a timestamp ($\text{time}_i$).

\noindent \textbf{Trajectory Transition Graph.} As shown in the left part of Fig.~\ref{fig:pre}, we utilize the widely adopted grid representation method~\cite{graphmm, mtrajrec, deeptrajrec} to split the city into $N_g = H \times W$ grids with side length $l_g$. This process results in each trajectory point $p$ falling into a specific grid denoted $g_{ij} = Grid(p)$, where $g_{ij}$ is the grid indexed by $(i,j)$ and $Grid(\cdot)$ can be considered as the function that maps trajectory points to grids. Therefore, a trajectory $\mathcal{T}$ can be represented by a sequence of grids with timestamps $\mathcal{T}=\langle g_1, \ldots, g_n \rangle$. By doing so, the relationships between different trajectories can be effectively captured from a global perspective. Specifically, we construct a directed trajectory transition graph $G_T=(V_T,E_T)$, where $V_T$ is the node set of grids and $E_T$ is the edge set representing trajectory transition. Specifically, as shown in the upper right part of Fig.~\ref{fig:pre}, for any trajectory $\mathcal{T}$ in trajectory set, if there is a trajectory point transition from $p_i \in \mathcal{T}$ to $p_{i+1} \in \mathcal{T}$ ($p_i$, $p_{i+1}$ are consecutive and $Grid(p_i)$, $Grid(p_{i+1})$ are not identical), then there is an edge form $Grid(p_i)$ to $Grid(p_{i+1})$. Moreover, the weight assigned to each edge reflects the count of such trajectory transitions observed across the trajectory set.

\noindent \textbf{Road Network.} A road network is a graph where the nodes represent locations, such as intersections, and the edges represent road segments connecting these locations.

\noindent \textbf{Link Connection Graph.} As shown in the lower right part of Fig.~\ref{fig:pre}, a directed link connection graph $G_R=(V_R,E_R)$ is used to model the road network, where each $u_i \in V_R$ is a road segment, and $e_{ij} = (u_i,u_j)$ represents road segment $u_i$ and $u_j$ are connected. To maintain consistency with the grid representation of trajectories, we choose the grid coordinates of the start and end points of road segments, along with their latitude and longitude as node features.

\subsection{Problem Formulation}

\noindent In the online scenario, we need to match newly arriving trajectory points at regular intervals. For each matching, we have the link connection graph $G_R$ which represents road network, the $k$ new incoming trajectory points $\mathcal{T}_{i:i+k}= \langle p_i, \ldots, p_{i+k-1} \rangle$ to match, the historical prefix trajectory $\mathcal{T}_{\vartriangleleft i} = \langle p_1, \ldots, p_{i-1} \rangle$ that is already available for previous matching, and the matched road $U_{\vartriangleleft i} = \langle u_1, \ldots, u_{i-1} \rangle$ of the historical prefix trajectory. Formally, we define the Online Map Matching problem and Online Markov Decision Process as follows.

\textit{\underline{Definition 1 (Online Map Matching).}} Given $G_R$, $\mathcal{T}_{i:i+k}$, $\mathcal{T}_{\vartriangleleft i}$ and $U_{\vartriangleleft i}$, the goal is to match $k$ road segments $U_{i:i+k} = \langle u_i, \ldots, u_{i+k-1} \rangle$ from their candidate sets $C_{i:i+k}$. The candidate sets are pre-computed based on spatial distance, which is a common practice in related works~\cite{HMM,AMM, mtrajrec}. The matching process is incremental and requires multiple steps to complete the matching of the entire trajectory $\mathcal{T}$.

\textit{\underline{Definition 2 (Online Markov Decision Process).}} The Markov Decision Process (MDP) provides a powerful framework for representing problems as a sequence of states and transitions, offering both effectiveness and efficiency. Inspired by this, we innovatively model the online map matching problem as an online MDP, meticulously tailored to capture the core aspects of the problem. Specifically, it is characterized by \textit{State}, \textit{Action}, \textit{Transition}, and \textit{Reward}.
\begin{itemize}[leftmargin=10.2pt]
\setlength{\itemsep}{0pt}
\setlength{\parsep}{0pt}
\setlength{\parskip}{0pt}
\item \textit{State}: The state $s_i$ of time step $i$ represents the real-time information of online map matching scenario (i.e., \textit{previously matched road segments} $U_i$, \textit{current trajectory points} $\mathcal{T}_i$, and \textit{candidate road segments} $C_i$), and the historical information of previous matches (i.e., \textit{historical information} of road $H_i^r$ and trajectory $H_i^t$). 

\item \textit{Action}: Action $a_i$ represents the selection of a road segment from the candidate road segments $C_i$. 
\item \textit{Transition}: In our modeling, the selected candidate road segments influence the update of historical information, which in turn affects the next state. Meanwhile, the previously matched road segments within the state are determined by the selected candidate road segments. This can be considered as state transitions in the context of map matching.
\item \textit{Reward}: Reward $r = R(s_i, a_i)$ denotes the reward received after taking action $a_i$ under state $s_i$ with reward function $R$, which is associated with the accuracy of the action.
\end{itemize}

By defining these components, the online map matching problem can be effectively modeled as an online MDP. In particular, previously defined $\mathcal{T}_{\vartriangleleft i}$, $U_{\vartriangleleft i}$, $\mathcal{T}_{i:i+k}$ and $C_{i:i+k}$ are effectively incorporate into \textit{State}, the \textit{Action} determines how to select $U_{i:i+k}$, the \textit{Transition} indicates the evolution of information for $\mathcal{T}_{\vartriangleleft i}$ and $U_{\vartriangleleft i}$ within \textit{State}, and the \textit{Reward} is the assessment of the matched results $U_{i:i+k}$. This approach enables the use of reinforcement learning algorithms to optimize decision-making for efficient and robust map matching in dynamic environments.

\section{Framework}
\label{sec:3}
\begin{figure}
  \centering  
  \includegraphics[width=0.7\linewidth]{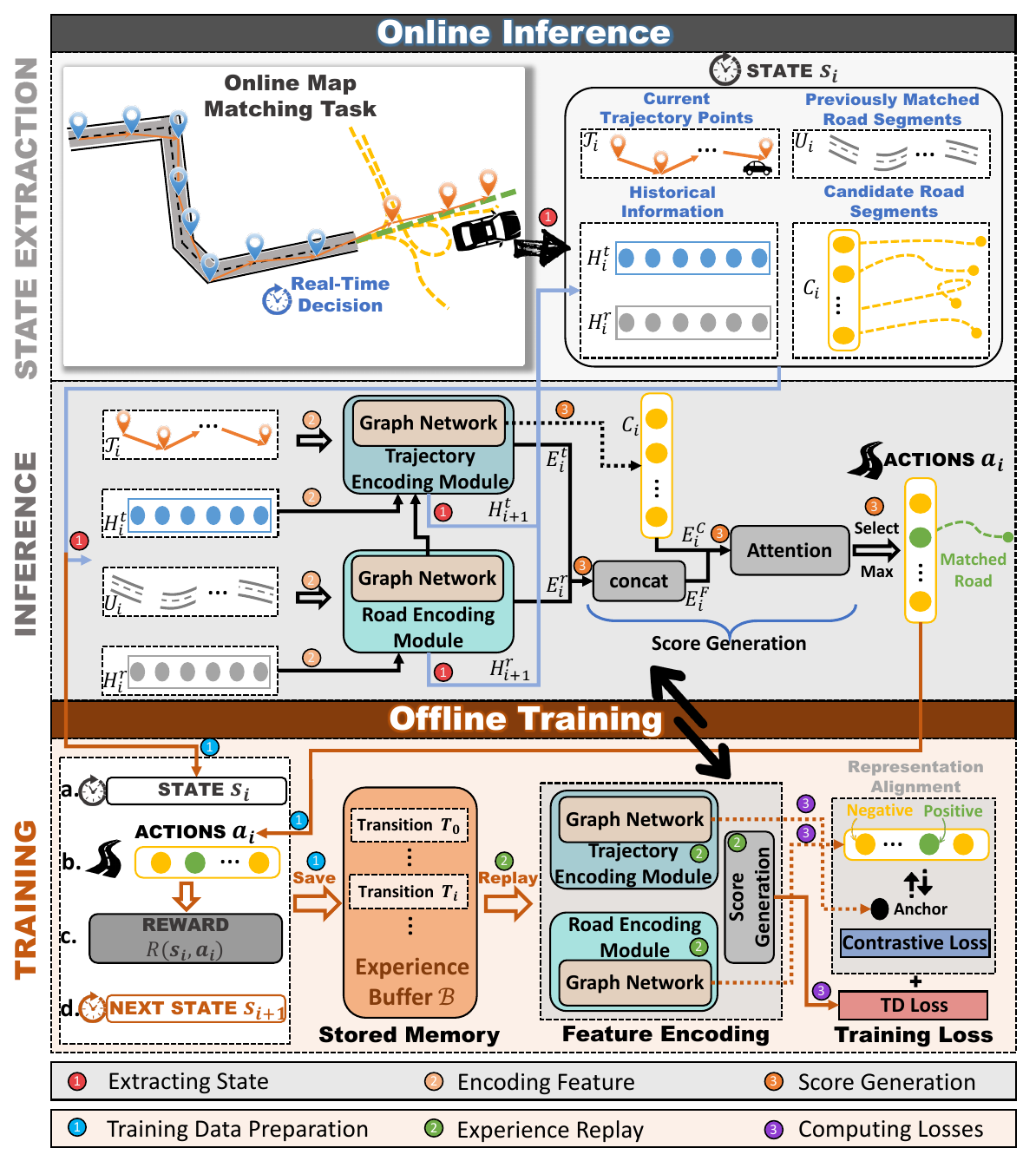}
  \vspace{-0.15in}
  \caption{The architecture of \modelName, which consists of two parts: Online Inference and Offline Training. }
  \vspace{-0.1in}
  \label{fig:overview}
\end{figure}

In this section, we outline the framework of our proposed \textbf{\modelName}. As shown in Fig.~\ref{fig:overview}, \textbf{\modelName} contains two stages: the online inference and the offline training.

\noindent \underline{\textit{\textbf{Online Inference.}}} At this stage, we first extract real-time information and historical information to form \textit{State}. Next, we design feature encoding modules to generate effective representations for trajectories and roads, and generate score for each candidate road segment to guide the matching decisions.

\textbf{1. Extracting State:} For a given time step $i$, we extract real-time information from the online map matching scenario, and preserve historical information from previous matches to construct state $s_i$. Specifically, the \textit{previously matched road segments} $U_i$, \textit{current trajectory points} $\mathcal{T}_i$, and \textit{candidate road segments} $C_i$ are incorporate as real-time information since they are basic elements of map matching. Moreover, to maximize the efficiency of information transfer during the online matching process and to prevent redundant computations, we retain historical information by preserving the hidden states from the encoding module of the preceding step, which contains \textit{historical information} of road $H_i^r$ and trajectory $H_i^t$. 

\textbf{2. Encoding Feature:} After the construction of the state $s_i$, we design the trajectory and road encoding module to obtain effective encodings of the corresponding information within the state. In the trajectory encoding module, we first construct the trajectory transition graph $G_T$ based on the set of trajectories, aiming to capture the inter-trajectory correlations to obtain informative trajectory representations. Subsequently, to bridge the interaction between representations of road segments and trajectory, we incorporate the representations of the corresponding road segment for the \textit{current trajectory point} $\mathcal{T}_i$ and apply a graph neural network to derive the trajectory representation. This is then combined with the historical information of trajectory $H_i^t$ to generate the trajectory encoding $E_i^t$ for time step $i$. In the design of the road encoding module, we initially employ the road network to construct the link connection graph $G_R$. Subsequently, analogous to the trajectory encoding module, we deploy a graph neural network to generate representations for road segments, and in conjunction with the historical information of road $H_i^r$, we obtain the road encoding $E_i^r$. Moreover, as mentioned before, the corresponding road segment representation for the \textit{current trajectory point} $\mathcal{T}_i$ is transmitted to the trajectory encoding module. Finally, the hidden states $H_{i+1}^r$ and $H_{i+1}^t$ produced by the encoding module are retained as the new historical information for the next time step $i+1$.

\textbf{3. Score Generation:} In this step, we first employ the graph neural network within the road encoding module to encode the candidate road segments $C_i$, generating candidates encoding $E_i^C$. Then, we concatenate the trajectory encoding $E_i^t$ and the road encoding $E_i^r$ to obtain fusion encoding $E_i^F$, and together with the candidate encoding $E_i^C$, we generate attention scores and select the maximum value as the matching decision action $a_i$.

\noindent \underline{\textit{\textbf{Offline Training.}}} At this stage, we first prepare the training data through multiple inferences and reward evaluations, saving it to the experience buffer. Next, we replay the experience through the feature encoding module to prepare for loss calculation. Finally, we compute different losses from the perspectives of temporal difference and contrastive learning.

\textbf{1. Training Data Preparation:} Given the action $a_i$ from the online inference matching decision and state $s_i$, we devise a meticulously reward evaluation function to generate reward $R(a_i,s_i)$. 
\textcolor{black}{Notably, the reward evaluation function thoroughly considers the complexity of online matching scenarios, performing a comprehensive assessment from a future-oriented perspective, which helps to minimize incorrect matches right from the start.}
After taking action $a_i$, we can obtain the next state $s_{i+1}$, which along with the state $s_i$, action $a_i$, and reward $R(a_i,s_i)$, are stored as a transition $T_i$ in the experience buffer. We repeat the aforementioned steps multiple times to prepare a collection of training data. This process will be conducted repeatedly to facilitate the model's continuous enhancement and refinement across various scenarios.

\textbf{2. Experience Replay:} After a certain number of online inferences and transition storages, we randomly select a batch of transitions from the experience buffer and replay them for training purposes. Specifically, these transitions are processed through our feature encoding module and score generation component, supplemented by double DQN techniques to prepare for loss calculation while reducing score overestimation.

\textbf{3. Computing Losses:} After the experience replay, we employ reinforcement learning strategies grounded in Temporal Difference (TD) learning principles. By leveraging the Temporal Difference (TD) Loss, the model can enhance the capability to estimate the q-value associated with action $a_i$ given the state $s_i$. Moreover, considering that the map matching problem necessitates effective encoding of both trajectories and road segments to integrate their information for accurate matching, we design a trajectory-road representation alignment module to align the representations in latent space. In particular, we select the representations of trajectory points as anchors, and take the representations of candidate road segments as positive and negative samples. Within the candidates, the ground truth and the remaining parts are considered as positive and negative, respectively. Finally, the contrastive loss is used together with the TD loss to train our model.

\section{Feature Encoding}
\label{sec:4}
\begin{figure}
  \centering  
  \includegraphics[width=0.7\linewidth]{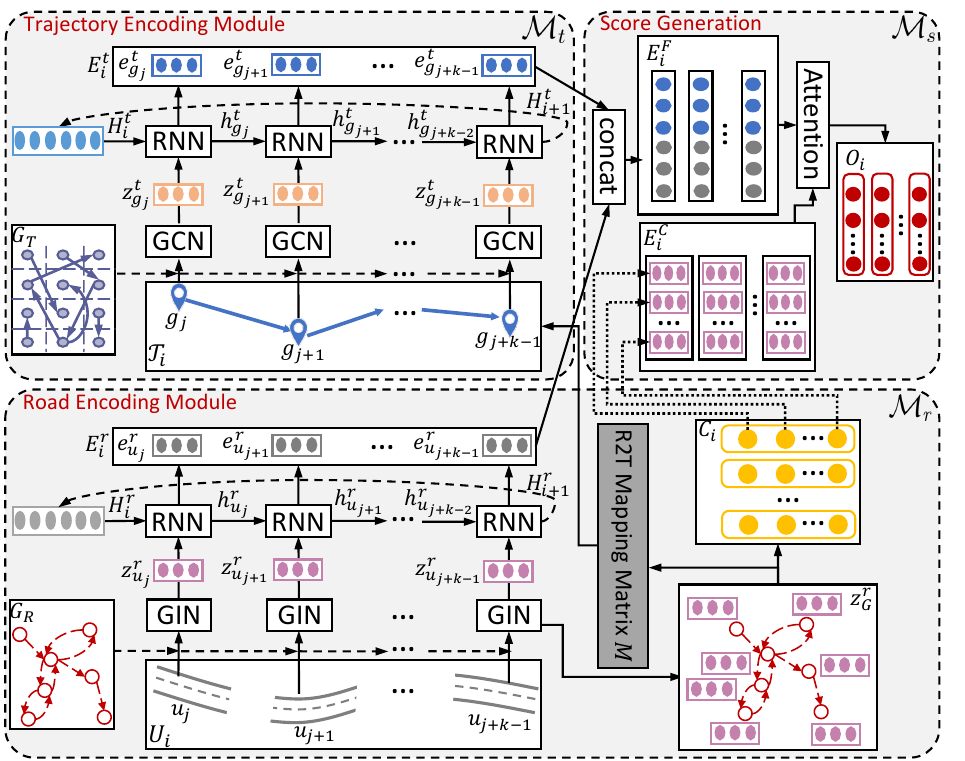}
  \vspace{-0.1in}
  \caption{The framework of feature encoding.}
  \vspace{-0.15in}
  \label{fig:sec4}
\end{figure}

As shown in Fig.~\ref{fig:sec4}, we design road encoding module $\mathcal{M}_r$, trajectory encoding module $\mathcal{M}_t$ and score generation module $\mathcal{M}_s$. In particular, we first elaborate on the details of deriving informative representations for trajectories and roads in Sec.~\ref{sec:4.1}, which is based on graph structures tailored to trajectories and roads. Then, we focus on efficiently capturing the sequence correlation in online map matching scenarios in Sec.~\ref{sec:4.2}. At last, we explain how to generate the score of each candidate in Sec.~\ref{sec:4.3}.

\subsection{Trajectory \& Road Representation}
\label{sec:4.1}
\noindent \textbf{Link Connection Graph Convolution.} The representation learning of road segments is crucial in the map matching problem. Considering that it can be naturally presented as a graph structure (i.e., the link connection graph $G_R$), we obtain its representation through a graph neural network. Specifically, considering the complex topological structure of road networks, motivated by~\cite{graphmm}, we adopt the Graph Isomorphism Network (GIN)~\cite{GIN}. GIN has the optimal ability to distinguish various graph structures, thus accurately representing the nuances of road networks. Its aggregation mechanism, which merges a node's features with those of its neighbors, is enhanced by a learnable parameter that precisely adjusts the balance between a node's own characteristics and its neighbors' influences. In our implementation, we adopt a lightweight version with a limited number of layers and a simple MLP, ensuring that this component is efficient for practical use. The update rule for feature $z_{u_j}^{(n)} \in \mathbb{R}^{d_r}$ of node $u_j \in V_R$ at the $n$-th layer can be formulated as follows:
\begin{equation}
    z_{u_j}^{(n)} = \text{MLP}^{(n)} ( (1 + \epsilon^{(n)}) \cdot z_{u_j}^{(n-1)} + \sum_{v \in \mathcal{N}(u_j)} z_v^{(n-1)})
\end{equation}
where $\text{MLP}^{(n)}$ denotes the multi-layer perceptron at the $n$-th layer, $\epsilon^{(n)}$ is a learnable parameter that adjusts the weight of a node's own features relative to its neighbors' features, $\mathcal{N}(u_j)$ represents the set of neighbor nodes of $u_j$, and $z_{u_j}^{(n-1)}$ and $z_{v}^{(n-1)}$ are the feature vectors of node $u_j$ and its neighbors, respectively. 
For previously matched road segments $U_i=U_{j:j+k}=\langle u_j, u_{j+1}, \ldots, u_{j+k-1} \rangle$, we use the feature $\langle z_{u_j}^{(n)}, z_{u_{j+1}}^{(n)}, \ldots, z_{u_{j+k-1}}^{(n)} \rangle$ after $n$-th GIN layer as the representations $\langle z_{u_j}^{r}, z_{u_{j+1}}^{r}, \ldots, z_{u_{j+k-1}}^{r} \rangle$ of road segments.

\noindent \textbf{Trajectory Transition Graph Construction.} The map matching problem is inherently graph-natured because the road network is represented as a graph composed of nodes (links) and edges (connections). Therefore, representing trajectories in a graphical form can also unify the representation of the two types of data, thereby alleviating the impact of their heterogeneity. Motivated by this, we further design trajectory transition graph $G_T = (V_T, E_T)$ to model the trajectories as a graph to address the heterogeneity between roads and trajectories.  Specifically, for trajectory set $\{\mathcal{T}_1, \mathcal{T}_2, \ldots, \mathcal{T}_n\}$ and trajectory point $g$ mapped to the grid, we construct the node set $V_T$ and $E_T$, which is defined as follows:
\begin{equation}
    V_T = \cup_{i=1}^{n} \{ g \mid g \in \mathcal{T}_i \}
\end{equation}
\begin{equation}
    E_T = \cup_{i=1}^{n} \{(g_j, g_{j+1}, w_{(g_j, g_{j+1})}) \mid g_j, g_{j+1} \in \mathcal{T}_i, g_j \neq g_{j+1} \}
\end{equation}
\begin{equation}
    w_{(g_j, g_{j+1})} = \sum_{i=1}^{n} \sum_{g_j, g_{j+1} \in \mathcal{T}_i} \mathbb{I}(g_j \neq g_{j+1})
\end{equation}
where $w$ is the weight of each edge, $\mathbb{I}(\cdot)$ is the indicator function, which takes a value of 1 if $g_j$ is different from $g_{j+1}$, and 0 otherwise. By doing so, the trajectory transition graph $G_T$ is able to model the relationships between trajectories from a global perspective of the trajectory set, where the edges and weights in the graph explicitly indicate the degree of correlation between the grids.

\noindent \textit{\textbf{Remark.}} In terms of complexity, the trajectory transition graph can be constructed with time and space complexity linear to the number of points in the trajectory set, both of which can be represented as $O(\sum_{i=1}^{n}|\mathcal{T}_i|)$, where $|\mathcal{T}_i|$ denotes the number of points in $\mathcal{T}_i$. This endows the trajectory transition graph with good scalability, enabling our model to be rapidly deployed in practical application scenarios with large amounts of trajectory data, and to incrementally update the graph structure during service provision.

\noindent \textbf{Trajectory Transition Graph Convolution.}
To enhance the consistency between trajectory and road representations and bridge their interaction, benefiting from our adoption of the same grid representation method for both trajectories and roads, we select the representations of road segments that intersect with the corresponding grid of the trajectory point as its initial features. Specifically, we construct a road-to-trajectory mapping matrix $M$, which distributes the road representations $z_r$ to various nodes of the trajectory transition graph $G_T$ based on whether the roads intersect with the nodes. It can be formulated as follows:
\begin{equation}
    M_{ij} = 
  \begin{cases} 
   1 & \text{if } g_i \cap u_j \neq \emptyset, \\
   0 & \text{otherwise}.
  \end{cases}
\end{equation}
where $\cap$ denotes the intersection of grid node and road segment. Then, for gird $g_i \in V_T$, its initial representation $z^{(0)}_{g_i} \in \mathbb{R}^{d_t}$ is:
\begin{equation}
    z^{(0)}_{g_i} = \frac{\sum_{j=1}^{N_g} M_{ij} \cdot z^r_{u_j}}{\sum_{j=1}^{N_g} M_{ij}}
\end{equation}
where $N_g$ is the number of nodes in $G_T$, $z^r_{u_j}$ denotes the representation of road segments $u_j$ of the global graph representation $z_G^r$. By doing this, the initial representations of the nodes in the trajectory transition graph $G_T$ sufficiently incorporate the associated road information. This strategy for linking road and trajectory representations lays a robust groundwork for the subsequent derivation of effective trajectory representations.

By definition, all transitions from grid $g_i$ to grid $g_j$ are aggregated into a single edge $e_{ij}$, resulting in a relatively sparse Trajectory Transition Graph. Given this sparsity, Graph Convolutional Networks (GCNs) are a suitable choice for encoding the graph, as they are computationally efficient on sparse graphs~\cite{GCN}. Using GCNs in this context allows for efficient encoding of the trajectory while aligning well with the graph’s structural characteristics. With the initial representations, we apply $n$ layers of Graph Convolutional Networks (GCN) to get the final representations. The update rule for feature $z_{g_i}^{(n)} \in \mathbb{R}^{d_t}$ of node $g_i \in V_T$ at the $n$-th layer can be formulated as follows:
\begin{equation}
    z_{g_i}^{(n)} = \sigma (\text{BN}^{(n)} (z_{g_i}^{(n-1)} \times {W}_a^{(n)} + \sum_{v \in \mathcal{N}(g_i)} \frac{1}{\sqrt{\hat{d}_{g_i} \hat{d}_v}} z_{v}^{(n-1)} \times {W}_b^{(n)} ))
\end{equation}
where $\sigma$ denotes the activation function ReLU, $\text{BN}^{(n)}$ denotes the batch normalization at the $n$-th layer, ${W}_a^{(n)} \in \mathbb{R}^{d_t \times d_t}$ and ${W}_b^{(n)} \in \mathbb{R}^{d_t \times d_t}$ are learnable weights, $\mathcal{N}(g_i)$ denotes the set of neighboring nodes of $g_i$, $\hat{d}_{g_i}$ and $\hat{d}_v$ are normalized degree terms for $g_i$ and $v$. For current
trajectory points $\mathcal{T}_i = \mathcal{T}_{j:j+k} = \langle g_j, g_{j+1}, \ldots, g_{j+k-1} \rangle$, we use the feature after $n$-th GCN layer as the representations $\langle z_{g_j}^{t}, z_{g_{j+1}}^{t}, \ldots, z_{g_{j+k-1}}^{t} \rangle$ of trajectory.

\subsection{Sequence Correlation Capture}
\label{sec:4.2}

In online map matching scenarios, we need to match newly generated trajectory points within a certain time interval. It's critical to take into account not only the historical information but also the intrinsic correlations present in the newly emerging sequence of trajectory points. In light of integrating them, a RNN model~\cite{RNN} is employed. Specifically, we leverage the sequential dependencies inherent in trajectory data by initializing the RNN with the hidden state derived from the previous trajectory points. This methodology facilitates a dynamic learning process, where the network progressively refines its understanding of the trajectory patterns, thereby improving the precision and efficiency of map matching process. 

For road representations $\langle z_{u_j}^{r}, z_{u_{j+1}}^{r}, \ldots, z_{u_{j+k-1}}^{r} \rangle$ and trajectory representations $\langle z_{g_j}^{t}, z_{g_{j+1}}^{t}, \ldots, z_{g_{j+k-1}}^{t} \rangle$, the update of hidden states and generation of output is denoted as follows:
\begin{equation}
h_{x_j} = \tanh(h_{x_{j-1}} \times {W}_{h} + b_{h} + z_{x_j} \times {W}_{z} + b_{z})
\end{equation}
\begin{equation}
e_{x_j} = h_{x_j} \times {W}_{e} + b_{e}
\end{equation}
where $x_j$ denotes road segment $u_j$ or current trajectory point $g_j$, $h_{x_j}$ and $e_{x_j}$ are hidden states and outputs, respectively. ${W}_h, {W}_z, {W}_e$ are the learnable weights, $b_h, b_z, b_e$ are the learnable bias. In particular, within our \modelName~framework's operation, $h_{u_{j-1}}^r$ and $h_{g_{j-1}}^t$ are included in the state $s_i$ of the online MDP we model, represented as historical information $H_i^r$ and $H_i^t$, respectively. Moreover, $h_{u_{j+k-1}}^r$ and $h_{g_{j+k-1}}^t$ are treated as the historical information $H_{i+1}^r$ and $H_{i+1}^t$ and saved for the next state $s_{i+1}$. After the sequence correlation capture of both trajectory and road segments, we get the trajectory embeddings $E_i^t = \langle e_{g_j}^t, e_{g_{j+1}}^t, \ldots, e_{g_{j+k-1}}^t \rangle$ and the road embeddings $E_i^r = \langle e_{u_j}^r, e_{u_{j+1}}^r, \ldots, e_{u_{j+k-1}}^r \rangle$ of state $i$.

\noindent \textit{\textbf{Remark.}} In online map matching, where immediate processing is essential, RNN~\cite{RNN} demonstrates advantages over GRU~\cite{GRU} and LSTM~\cite{LSTM} due to its simpler structure and faster computational capabilities. Additionally, compared to Transformer~\cite{transformer}, which lacks sequential hidden state transfer, RNN maintains a continuous context flow that aligns well with MDP modeling, where the current state encapsulates all necessary past information, reflecting the Markov property essential for sequential data processing. In this scenario, the relevance of long-term historical data to current matching is minimal, as distant trajectory information has little impact on immediate decisions. This context is congruent with RNN's operational strengths and mitigates their limitation of handling long-term dependencies~\cite{RNNsurvey}, making their efficient processing of short-term, relevant sequences particularly well-suited for the quick-paced, real-time requirements of online map matching.

\subsection{Score Generation}
\label{sec:4.3}
Based on considerations of consistency, for the candidate road segments $C_i \in \mathbb{R}^{k \times n_c}$ within state $i$, we use the GIN in the road encoding module to obtain their embeddings, where $k$ is the matching interval and $n_c$ is the number of candidates. Specifically, we select the representations of the corresponding road segments from the global graph representation $z_G^r$ as the candidates embedding $E_i^C \in \mathbb{R}^{k \times n_c \times d_r}$. Subsequently, we concatenate the trajectory embedding $E_i^t \in \mathbb{R}^{k \times d_t}$ and road embedding $E_i^r \in \mathbb{R}^{k \times d_r}$ of state $i$ to obtain a fused embedding $E_i^F \in \mathbb{R}^{k \times (d_r+d_t)}$.
Next, we apply an attention mechanism to calculate attention scores $O_i \in \mathbb{R}^{k \times n_c}$ based on the fused embedding $E_i^F$ and candidate road segments embedding $E_i^C$, serving as the scores for each candidate road segment. In particular, we transform $E_i^F$ and $E_i^C$ respectively and use them as the query $Q = \tanh(E_i^F \times W_F+ b_F)$ and key $K = \tanh(E_i^C \times W_C + b_C)$ in the attention mechanism to generate the score $O_i = Q \cdot K^\top$, where $W_F \in \mathbb{R}^{(d_r+d_t) \times d_a}$ and $W_C \in \mathbb{R}^{d_r \times d_a}$ are the learnable weights, $b_F \in \mathbb{R}^{d_a}$ and $b_C \in \mathbb{R}^{d_a}$ are the learnable bias.

\section{Model Learning}
\label{sec:5}
In this section, we first focus on the trajectory-road representation alignment module in Sec~\ref{sec:5.1}. Next, we describe the reward evaluation in Sec.~\ref{sec:5.2}. Finally, in Sec.~\ref{sec:5.3}, we provide the details of the entire training process of the model and complexity analysis.
\subsection{Trajectory-Road Representation Alignment}
\label{sec:5.1}
The core challenge of map matching lies in accurately aligning trajectories to the corresponding road network. To address this, we design a trajectory-road representation alignment module that fundamentally enhances the ability of our model to discern and align trajectory data with road segments.


Based on the effective trajectory and road representations detailed in Sec.~\ref{sec:4.1}, we first obtain representations for trajectory requiring matching and its corresponding candidate road segments. Specifically, we use the trajectory representation $z_{\mathcal{T}_i}^t$ for trajectory $\mathcal{T}_i$ as anchor $\mathcal{A}_{i} \in \mathbb{R}^{k \times d_t}$, and the embeddings for candidates $E_i^C \in \mathbb{R}^{k \times n_c \times d_r}$ as sample set. Among the sample set, we select the representation for ground truth as positive samples $\mathcal{P}_{i} \in \mathbb{R}^{k \times d_r}$ and the representation for remaining parts as negative samples $\mathcal{N}_{i} \in \mathbb{R}^{k \times (n_c-1) \times d_r}$. The process can be formulated as follows:
\begin{equation}
    \{\mathcal{A}_{i}, \mathcal{P}_{i}, \mathcal{N}_{i}\} = \left\{
    \begin{aligned}
        &\mathcal{A}_{i} = z_{\mathcal{T}_i}^t \\
        &\mathcal{P}_{i} = \{z_{u_j}^r \in E_i^C \mid u_j \in y_{i}\} \\
        &\mathcal{N}_{i} = \{z_{u_j}^r \in E_i^C \mid u_j \notin y_{i}\}
    \end{aligned}
    \right.
\end{equation}
where $z_{\mathcal{T}_i}^t \in \mathbb{R}^{k \times d_t}$ is the representation for trajectory $\mathcal{T}_i$ generated by the trajectory encoding module, $z_{u_j}^r \in \mathbb{R}^{d_r}$ denotes the representation of candidate road segment $u_j$ generated by the road encoding module, and $y_{i}$ denotes the ground truth road segments for trajectory $\mathcal{T}_i$.

Then, we introduce the InfoNCE loss~\cite{infonce} as an unsupervised alignment loss $\mathcal{L}_a$ to bring the anchor and positive samples closer together while distancing them from negative samples.
\begin{equation}
\mathcal{L}_a = -\frac{1}{N} \sum_{i=1}^{N} \log \frac{\exp\left(\frac{\mathcal{A}_i^\top \mathcal{P}_i}{\tau}\right)}{\exp\left(\frac{\mathcal{A}_i^\top \mathcal{P}_i}{\tau}\right) + \sum_{\mathbf{z}_{u_j}^r \in \mathcal{N}_i} \exp\left(\frac{\mathcal{A}_i^\top \mathbf{z}_{u_j}^r}{\tau}\right)}
\end{equation}
where $N$ is the number of anchor-sample pairs, and $\tau$ scales the softmax function. By doing so, the robustness of the representations for trajectories and roads is enhanced, enabling effective integration to achieve better matching performance.

\subsection{Reward Evaluation}
\label{sec:5.2}
After the score generation for each candidate road segment in Sec.~\ref{sec:4.3}, we get the scores $O_i \in \mathbb{R}^{k \times n_c}$ for $k$ time step of state $s_i$. Then, we select the maximum value in $O_i$ as the matching result for the current state $s_i$, which also serves as the action $a_i$:
\begin{equation}
    a_i^{(n)} = u_{k^*}, \text{ where } k^* = \arg \max_{j \in \{1, 2, \ldots, n_c\}} O_j^{(n)}
\end{equation}
where $a_i^{(n)}$ denotes the time step $n$ of action for state $i$, $\arg \max$ denotes the argument of the maximum, which is used to find the index $j$ that maximizes the value of $O_j^{(n)}$.

Upon obtaining the action $a_i$ corresponding to the current state $s_i$, in alignment with the OMDP modeling, we devise a reward evaluation mechanism to ascertain the reward. Specifically, we consider the reward from four aspects: \textit{Accuracy}, \textit{Consecutive Success}, \textit{Detour Penalty}, and \textit{Road Connectivity}.

\textbf{Accuracy.} The primary goal of map matching is to align the observed trajectory accurately with the corresponding road segments. The matching accuracy reward is designed to directly incentivize the model to select road segments that best fit the GPS points, which is formulated as follows:
\begin{equation}
   r_{ac} = \begin{cases} 
    1 & \text{if } a_i^{(n)} = y_{s_i}^{(n)} \\
    -1 & \text{otherwise}
\end{cases}
\end{equation}
where $a_i^{(n)}$ denotes the action of time step $n$, and $y_{s_i}^{(n)}$ is the ground truth road segments for state $s_i$ at time step $n$.

\textbf{Consecutive Success.} In real-world scenarios, trajectory points are collected sequentially as a vehicle moves. Consecutive success reward is given for consecutive successful matches along a trajectory, which motivates the model to not only focus on individual, isolated matchings but also to ensure that these matchings are logically consistent over time. This reward helps model to maintain continuity in the sequence of matched road segments, thereby improving overall accuracy from a future-oriented perspective.
\begin{equation}
    r_{cs} = \alpha \cdot \mathbb{I}(a_i^{(n)}= y_{s_i}^{(n)} \text{ and } m \geq \theta)
\end{equation}
where $\mathbb{I}(\cdot)$ is the indicator function, which takes a value of 1 if the condition is satisfied and 0 otherwise. $m$ denotes the current number of consecutive success, $\theta$ is the predefined threshold, and $\alpha$ is the reward value.

\textbf{Detour Penalty.} Unnecessary detours are a significant issue in map matching, often caused by GPS errors where consecutive trajectory points erroneously match upstream of the previous point~\cite{mmsurvey,LHMM,DMM,AMM}. This issue severely affects the accuracy of matches, leading to abnormal and circuitous results in the matched routes. Motivated by this, the detour penalty is implemented to discourage the model from choosing circuitous or uncharacteristically long paths that deviate significantly from the most direct route between consecutive GPS points.
\begin{equation}
    r_{dp} = -\beta \cdot \mathbb{I}(a_i^{(n)} \neq y_{s_i}^{(n)} \text{ and } a_i^{(n)} \in H)
\end{equation}

where $H$ is a historical matching queue that stores recent matching results, and $\beta$ is the penalty value.

\textbf{Road Connectivity.} Ensuring that the selected road segments are interconnected is crucial for generating viable routes in map matching. A road connectivity reward is designed to promote the selection of navigable paths and help the model understand that accurate matching is constrained by the connectivity between roads. We specifically consider the shortest path distance between the two road segments within the link connection graph $G_R$ as the degree of connectivity. The rationale behind this is that a shorter shortest path distance signifies a stronger connection between the segments.
\begin{equation}
    r_{rc} = \gamma \cdot \frac{1}{\delta(a_i^{(n)}, a_i^{(n-1)})}
\end{equation}
where $\delta(\cdot)$ represents the degree of connectivity of current road and previous road, and $\gamma$ is the reward value. Note that in this paper, we use a fixed road network, meaning we do not consider dynamic changes such as temporary road closures, which is a common practice in existing works~\cite{mtrajrec,l2mm,graphmm, AMM}. However, we can address this issue to some extent by introducing a dynamic connectivity discrimination function $\delta^d(\cdot)$. This is difficult for other existing methods, as they do not explicitly account for the impact of road connectivity on matching results, making it hard for them to extend this capability.

Finally, we can evaluate the overall reward as follows:
\begin{equation}
    r = r_{ac} + r_{cs} + r_{dp} + r_{rc}
\end{equation}

\begin{figure}[!t]
\begin{algorithm}[H]
    \small 
    \SetKwInOut{Input}{Input}
    \SetKwInOut{Output}{Output}
    \Input{training inputs $\mathbf{T}, \mathbf{C}, \mathbf{y}$, network module $\mathcal{M}_t$, $\mathcal{M}_r$, $\mathcal{M}_s$, trajectory transition graph $G_T$, link connection graph $G_R$, experience buffer $\mathcal{B}$, matching steps $k$, learning rate $lr$, training epochs $ep$, batch size $bs$, loss weight $\lambda$, update interval $t$, discount factor $\gamma$.}
    \Output{parameters $\theta_t$, $\theta_r$, $\theta_s$, for the three parts $\mathcal{M}_t$, $\mathcal{M}_r$, $\mathcal{M}_s$.}
    \caption{Model Learning for \textbf{\modelName}}
    \label{alg:ml}
    initialize $\theta_t, \theta_r, \theta_s$ with normal distribution\;
    \For{$m \gets 1 \ldots ep$}{
        iterations $I = \lfloor \frac{|\mathbf{T}|}{bs} \rfloor$\;
        \For{$n \gets 1 \ldots I$}{
            fetch the batch $\{\mathcal{T}, C, y\}$ from \{$\mathbf{T}, \mathbf{C}, \mathbf{y}$\}\;
            inference times $IT = \lfloor \frac{|\mathcal{T}|}{k} \rfloor$\;
            \For{$i \gets 1 \ldots IT$}{
                $s_i \gets \{\mathcal{T}_i, U_i, H_i^t, H_i^r, C_i\}$\;
                $a_i, r_i \gets ExperienceInference(s_i, G_T, G_R)$\;
                $s_{i+1} \gets \{\mathcal{T}_{i+1}, U_{i+1}, H_{i+1}^t, H_{i+1}^r, C_{i+1}\}$\;
                construct transition $T_i = \{s_i, a_i, r_i, s_{i+1}\}$\;
                store $T_i$ into $\mathcal{B}$ based on First In First Out\;
            }
            $\theta_{t}, \theta_{r}, \theta_{s} \gets ModelTrain(\mathcal{B}, G_T, G_R, lr, \lambda, \gamma)$\;
            update the main network with $\theta_{t}, \theta_{r}, \theta_{s}$\;
            \If{$n\mod t = 0$}{
                using parameters of the main network to update the target network\;
            } 
        }
    }
\end{algorithm}

\begin{algorithm}[H]
    \small 
    \SetAlgorithmName{Function}{function}{List of Functions}
    \renewcommand{\thealgocf}{}  
    \SetKwInOut{Input}{Input}
    \Input{current state $s_i$, trajectory transition graph $G_T$, link connection graph $G_R$.}
    \caption{ExperienceInference}
    \label{alg:ei}
    $\mathcal{T}_i, U_i, H_i^t, H_i^r, C_i \gets s_i$\;
    $H_{i+1}^t, E_i^t \gets \mathcal{M}_t(\mathcal{T}_i, H_i^t, G_T)$\;
    $H_{i+1}^r, E_i^r \gets \mathcal{M}_r(U_i, H_i^r, G_R)$\;
    $E_i^C \gets \mathcal{M}_r(C_i, G_R)$\;
    $O_i \gets \mathcal{M}_s(E_i^t, E_i^r, E_i^C)$\;
    $a_i \gets$ using Equation 12 with $O_i$\;
    $r_i \gets$ using Equations 13-17 with $a_i$ and $s_i$\;
    return $a_i, r_i$\;
\end{algorithm}

\begin{algorithm}[H]
    \small 
    \SetAlgorithmName{Function}{function}{List of Functions}
    \renewcommand{\thealgocf}{}  
    \SetKwInOut{Input}{Input}
    \Input{experience buffer $\mathcal{B}$, trajectory transition graph $G_T$, link connection graph $G_R$, learning rate $lr$, loss weight $\lambda$, discount factor $\gamma$.}
    \caption{ModelTrain}
    random sample transitions $T$ from $\mathcal{B}$\;
    training iterations $TI = |T|$\;
    \For{$i \gets 1 \ldots TI$}{
        $s_i, a_i, r_i, s_{i+1} \gets T_i$\;
        $q \gets Q(s_i, a_i; \theta_{main})$\;
        $a_{i+1}^* \gets \arg\max_{a_{i+1}} Q(s_{i+1}, a_{i+1}; \theta_{main})$\;
        $q_{target} \gets r_i + \gamma \cdot Q(s_{i+1}, a_{i+1}^*; \theta_{target})$\;
        $\mathcal{A}_i, \mathcal{P}_i, \mathcal{N}_i \gets $ using Equation 10\;
        $\mathcal{L} \gets$ using Equations 11, 18-19 and $\lambda$\;
    }
    return $\theta_{t}, \theta_{r}, \theta_{s}$;
\end{algorithm}  
\vspace{-0.2in}
\end{figure}

\subsection{Model Training}
\label{sec:5.3}
To leverage the effective sequential decision-making capabilities of reinforcement learning, as shown in Algorithm~\ref{alg:ml}, we develop a novel model learning algorithm that enhances the model's adaptability and robustness across diverse scenarios. For each epoch, we first fetch a batch of trajectories, candidates and ground truths $\{\mathcal{T}, C, y\}$ from whole set of trajectories, candidates and ground truths $\{\mathbf{T}, \mathbf{C}, \mathbf{y}\}$ (line 5 of Algorithm~\ref{alg:ml}). Then, the systematic accumulation of transitions through \textit{Experience Inference} is conducted, which are then stored in an experience buffer (lines 7-12 of Algorithm~\ref{alg:ml}). Next, the model continuously samples from these stored transitions, enabling robust \textit{Model Train} that incorporates a wide range of experiences (lines 13-16 of Algorithm~\ref{alg:ml}). Through this alternating process, this learning approach enhances the model’s generalization ability while reducing the risk of overfitting, achieving high adaptability and robustness, making it suitable for real-world applications.

\noindent \textbf{Experience Inference.} The process begins by using the model to encode features and generate the scores (lines 1-5 of Function: ExperienceInference), which are detailed in Sec~\ref{sec:4}. Then, we select the action $a_i$ and evaluate the reward for action $a_i$ under state $s_i$ to obtain the reward $r_i$ that encompasses multiple aspects (lines 6-7 of Function: ExperienceInference). Together with the next state $s_{i+1}$, we construct a transition $T_i = \{s_i, a_i, r_i, s_{i+1}\}$ as a piece of experience and store it in the experience buffer $\mathcal{B}$. We refer to this process as the \textit{experience inference} process, and execute it multiple times to accumulate a sufficient amount of transitions.

\noindent \textbf{Model Train.} After a certain number of experience inferences, the model randomly samples a batch of transitions from the experience buffer $\mathcal{B}$ for experience replay. Specifically, we employ a deep Q-learning method, which learns the value of an action in a particular state, intending to maximize the total reward received~\cite{dqn}. Moreover, to reduce the overestimation of action values that often occurs in standard Q-learning, we apply Double DQN~\cite{double_dqn} to increase the stability of our training. In particular, we use two networks with identical structure that are composed of $\mathcal{M}_t$, $\mathcal{M}_r$, and $\mathcal{M}_s$, namely the main network and the target network. Note that we only train the main network, and use its parameters to update the target network periodically (lines 15-16 of Algorithm~\ref{alg:ml}). We first use the main network to estimate the q-value $q$ (line 5 of Function: ModelTrain). Then, we use the main network to select the action $a_{i+1}^*$ with the max q-value (line 6 of Function: ModelTrain). Next, we use the target network to estimate the Q-value of taking action $a_{i+1}^*$ under next state $s_{i+1}$, multiply it by the discount factor $\gamma$, and add it with reward $r_i$ to compute the target q-value $q_{target}$ (line 7 of Function: ModelTrain). Finally, we apply the Huber Loss~\cite{huber} as the Temporal Difference (TD) loss for Q-learning, which is formulated as follows:
\begin{equation}
    \mathcal{L}_{td} = 
    \begin{cases} 
    0.5 \times (q - q_{\text{target}})^2, & \text{if } |q - q_{\text{target}}| < 1 \\
    |q - q_{\text{target}}| - 0.5, & \text{otherwise}
    \end{cases}
\end{equation}
where $q$ is the estimated q-value and $q_{target}$ is the target q-value. Combined with the alignment loss $\mathcal{L}_a$ and corresponding weights $\lambda$, we can compute the overall learning objective as follows:
\begin{equation}
    \mathcal{L} = \mathcal{L}_{td} + \lambda \cdot \mathcal{L}_a
\end{equation}

\begin{table}[!t]
\centering
\vspace{-0.1in}
\caption{Complexity Analysis of the Main Methods}
\vspace{-0.1in}
\label{tab:comp}
\scalebox{0.9}{\begin{tabular}{c|c|c}
\hline
 & \textbf{\#Params} & \textbf{Complexity} \\
\hline
\textbf{MDP} & $O(n_R \times c)$ & $O( n_T\times \frac{l_T}{a} \times K \times n_R \times c)$\\
\hline
\textbf{DNN} & $O(n_R\times d)$ & $O(n_T\times \frac{l_T^2}{2a}\times n_R \times d)$\\
\hline
\textbf{HMM} & $O(n_R \times c)$ & $O(n_T \times l_T \times c \times M_{\max})$ \\
\hline
\textbf{\modelName} & $O(d^2)$ & $O(n_T\times \frac{l_T}{a}\times d^2)$ \\
\hline
\end{tabular}}
\vspace{-0.15in}
\end{table}
\noindent \textbf{Complexity Analysis.} We compare \textbf{\modelName}~with \textbf{MDP-based}~\cite{MDPMM}, \textbf{DNN-based}~\cite{mtrajrec, l2mm, graphmm} offline methods, and \textbf{HMM-based}~\cite{HMM, FMM, AMM} online method in terms of complexity. We analyze the complexity of parameters and computation for the four types of methods, with the results shown in Table~\ref{tab:comp}. Specifically, we model the problem input as follows. The road network contains $n_R$ road segments. The number of trajectories is $n_T$, the longest trajectory length is $l_T$, and each time interval has $a$ new trajectory points. For each trajectory point, the maximum candidates road segments is $c$. The dimension of all hidden representations is $d$.

For \textbf{MDP} methods, the parameters mainly consist of transition probabilities for each state-action pair, which has the complexity of $O(n_R \times c)$. The computational complexity primarily arises from calculating the value function for each state-action pair during the iterative process (e.g., value iteration). Assuming the number of iterations required for convergence is $K$, the time complexity for a single computation is $O(K \times n_R \times c)$. In the online scenario, the method needs to recompute the MDP multiple times, with the trajectory length increasing each time. The cumulative computational complexity becomes $O( n_T\times \frac{l_T}{a} \times K \times n_R \times c)$, where $\frac{l_T}{a}$ represents the number of calculations for each trajectory.
For \textbf{DNN} methods, the FC layer and the road segment embedding layer have $O(n_R \times d)$ parameters, where $n_R >> d$, making these parameters significantly greater than the $O(d^2)$ parameters of other layers. This results in an overall parameter complexity of $O(n_R \times d)$. In the online scenario, the model needs to recompute over increasingly longer trajectories, leading to an accumulated computational complexity of $O(n_T\times \frac{l_T^2}{2a}\times n_R \times d)$, where the term $\frac{l_T^2}{2a}$ comes from summing the lengths of all steps for a trajectory, which forms an arithmetic series.
For \textbf{HMM} methods, the parameter complexity is $O(n_R \times c)$, which comes from emission probabilities and transition probabilities between states. Assuming the maximum number of transitions per candidate is $M_{\max}$, at each time step, HMM methods can perform incremental online matching, with a total time complexity of $O(n_T\times l_T \times c \times M_{\max})$. 
For our \textbf{\modelName}~method, it does not require embedding encoding of $n_R$ road segments, therefore the parameter complexity is $O(d^2)$. Since our method efficiently retains historical information and can match $a$ trajectory points at a time, the time complexity is reduced to $O(n_T \times \frac{l_T}{a} \times d^2)$.

\textbf{\textit{Remark:}} The complexity of \textbf{MDP} and \textbf{DNN} method has the large term $n_R$, $K$ and the quadratic term $\frac{l_T^2}{2a}$, making them inefficient because they redundantly process the entire trajectory. In contrast, in our method, $d$ is much smaller than $n_R$, and $\frac{l_T}{a}$ achieves optimal complexity with respect to the trajectory length term. Additionally, since $M_{\max}$ is relatively small compared to $n_R$, the complexity of \textbf{HMM} methods is relatively acceptable. However, it's term $l_T$ still exceeds our method’s  $\frac{l_T}{a}$.
Therefore, our \textbf{\modelName} achieves superior efficiency by eliminating redundant computations, making it well-suited for real-world applications.

\begin{table}[!t]
\centering
\caption{Statistics of Datasets}
\vspace{-0.15in}
\label{tab:Datasets}
\scalebox{0.85}{\begin{tabular}{|c|c|c|c|}
\hline
 & \textbf{Beijing} & \textbf{Porto} & \textbf{Chengdu}\\
\hline
\textbf{Time Span} & 03/01-06/23, 2022 & 01/07-06/30, 2014 & 11/01-11/30, 2016\\
\textbf{Cover Area}  & 8.69km × 7.67km & 5.86km × 3.34km & 8.32km × 8.36km\\
\textbf{Road Segments} & 8.5K & 4.2K & 6.5K\\
\textbf{Trajectories} & 10K & 129.5K & 898.1k\\
\textbf{Sampling frequency} & 15s & 15s & 15s\\
\textbf{Average Error} & 23.9m & 7.8m & 12.7m\\
\hline
\end{tabular}}
\vspace{-0.2in}
\end{table}

\section{Experiments}
\label{sec:6}
\subsection{Experimental Setup}
\noindent \textbf{Datasets.} As shown in Table~\ref{tab:Datasets}, we conduct a series of experiments based on three real-world datasets.

\noindent (1) \textbf{Beijing}~\cite{graphmm}: This dataset is collected from the real-world application Tencent Maps\footnote{https://map.qq.com} and is a large-scale dataset in terms of the number of road segments.

\noindent (2) \textbf{Porto}~\cite{kaggle}: This dataset of trajectories is open-sourced on Kaggle. The road network of this dataset is extracted from OpenStreetMap\footnote{https://www.openstreetmap.org \label{openstreet}}. 

\noindent (3) \textbf{Chengdu}~\cite{didi}: This dataset is open-sourced by DiDi to support related research. The road network of this dataset is extracted from OpenStreetMap\footref{openstreet}. This dataset contains a large number of trajectories, making it suitable for evaluating the scalability of various methods in large-scale data scenarios.

\noindent (4) \textbf{Data Labels}: In the map matching problem, data labels typically come from two sources: real-world collection or HMM annotation under high sampling rates. For Beijing, thanks to work by Tencent Maps, the labels are entirely real, which is rare in the realm of trajectory data. For Porto and Chengdu, we use HMM to annotate the data under high sampling rates (15s). This is a common practice in the related works~\cite{LHMM, FL-AMM, l2mm}, as the effectiveness of HMM for high-sampling-rate data is already a consensus. In practical scenarios, high-sampling-rate data can be used to generate high-quality labels, serving as reliable training data to support the training of deep learning methods. Once the model is trained, deploying it in low-sampling-rate scenarios is more cost-effective, as it avoids the communication, storage, and computational overhead associated with high-sampling-rate data. Therefore, different sampling rates can be utilized at different stages, ensuring the quality of training data while balancing cost and performance in practical deployment scenarios.

\noindent (5) \textbf{Training, Validation and Test}: We divide a dataset into training, validation and test data with the splitting rate of 7:2:1. We set the sampling rate of the dataset as 50\%, 25\%, and 12.5\%, corresponding to the time intervals of 30s, 60s, and 120s. In particular, it makes the trajectory points sparse and therefore challenging for map matching algorithms. We set the matching interval to 120 seconds, which is a common processing interval in practical applications~\cite{graphmm}, thus the number of matching steps $k$ for sampling rates of 50\%, 25\%, and 12.5\% is 4, 2 and 1. The reward value $\alpha, \beta, \gamma$ is set to 0.01, 0.05 and 0.02, respectively. Moreover, we set 10 as the number of candidates $n_c$, thereby selecting 10 candidate road segments closest to each trajectory point. For Beijing, Porto and Chengdu, this allows our model to consider nearly all possible road segments within distances of 160 meters, 140 meters and 175 meters from the trajectory points, respectively. This is already significantly larger than the Average Error of 23.9 meters for Beijing, 7.8 meters for Porto and 12.7 meters for Chengdu.

\noindent \textbf{Baseline methods.}
We compare the following baselines:
\begin{itemize}[leftmargin=10pt]
    \item \textbf{MDP}: We employ the value iteration algorithm based on Markov Decision Process (MDP) to perform map matching.
    \item \textbf{HMM}~\cite{HMM}: Hidden Markov Model (HMM) is a classical probabilistic framework that models potential road segments as states and GPS data as observations.
    \item \textbf{FMM}~\cite{FMM}: This is a HMM-based method store shortest paths in a precomputed table with quick hash table searches, which achieves high processing speed.
    \item \textbf{AMM}~\cite{AMM}: This is the current state-of-the-art method for online map matching. It designs a collaborative evaluation model with a retrospective correction mechanism.
    \item \textbf{MTrajRec}~\cite{mtrajrec}: This is a method that innovatively integrates trajectory recovery and map matching for urban applications, using a sequence-to-sequence learning model to enhance low-sampling-rate GPS data. 
    \item \textbf{L2MM}~\cite{l2mm}: This method generates representations of low-quality trajectories through high-frequency trajectory enhancement and data distribution augmentation, and incorporates mobility patterns into the map matching task.
    \item \textbf{GraphMM}~\cite{graphmm}: This is the current state-of-the-art method for offline map matching, which proposes a graph-based solution to capture the correlations of trajectories and road segments.
\end{itemize}
For the offline map matching methods, we adjust them to online form to suit online scenarios. Specifically, because offline methods lack the transmission of historical information of the current matching trajectory, we perform incremental multiple matches to meet the requirements of online scenarios. \textcolor{black}{Note that it's fair for them, as these offline methods obtain all the currently available trajectories each time a match is made, which is consistent with the setting of offline matching problem they aim to solve.}

\noindent \textbf{Evaluation Metrics.} We evaluate our proposed methods and baseline methods based on two metrics: AccT (Trajectory-level Accuracy) and LCSR (Longest Common Subsequence Ratio). Compared to traditional metrics such as precision and recall, these two metrics emphasize the sequential nature of matching, which is crucial for online scenarios and adopted by real-world applications~\cite{graphmm}. 
Specifically, for each trajectory $\mathcal{T}_i$ in trajectory set $T$, suppose the ground truth and matched result of model of road segments is represented as $\langle u_{1}, \cdots, u_{l_i} \rangle$ and $\langle \hat{u}_{1}, \cdots, \hat{u}_{l_i} \rangle$, the accuracy of trajectory set $T$ is computed as follows: $AccT(T) = \frac{1}{n}\sum_{i=1}^{n}\frac{\sum_{j=1}^{l_i}\mathbb{I}(u_j=\hat{u}_j)}{l_i}$, where $n$ is number of trajectories in trajectory set $T$, $l_i$ is the length of trajectory $\mathcal{T}_i$, and $\mathbb{I}(\cdot)$ is the indicator function, which takes a value of 1 if the matched road segment is identical to the ground truth, and 0 otherwise.
The LCSR metric characterizes the similarity between two sequences and is well-suited for evaluating tasks like online map matching that emphasize the order of sequences. In particular, for trajectory set $T$, it is compute as follows: $LCSR(T) = \frac{1}{n}\sum_{i=1}^{n}\frac{LCS(\mathcal{T}_i, \hat{\mathcal{T}}_i)}{l_i}$, where $LCS(\cdot)$ is longest common subsequence function.

\noindent \textbf{Experimental Settings.} All deep learning methods were implemented with PyTorch 2.0.1 and Python 3.11.5, and trained with a Tesla V100 GPU. All rule-based methods were implemented with python 3.11.5, and tested with a Intel Xeon Gold 6148 CPU @ 2.40GHz. The platform ran on Ubuntu 18.04 OS. In addition, we used Adam~\cite{adam} as the optimization method with the mini-batch size of 512. The learning rate is set as 0.001, and the training epoch is set as 200, and an early stopping mechanism is adopted.

\subsection{Setting of Model's Hyper-parameters}
\begin{figure}
  \centering  
  \includegraphics[width=0.7\linewidth]{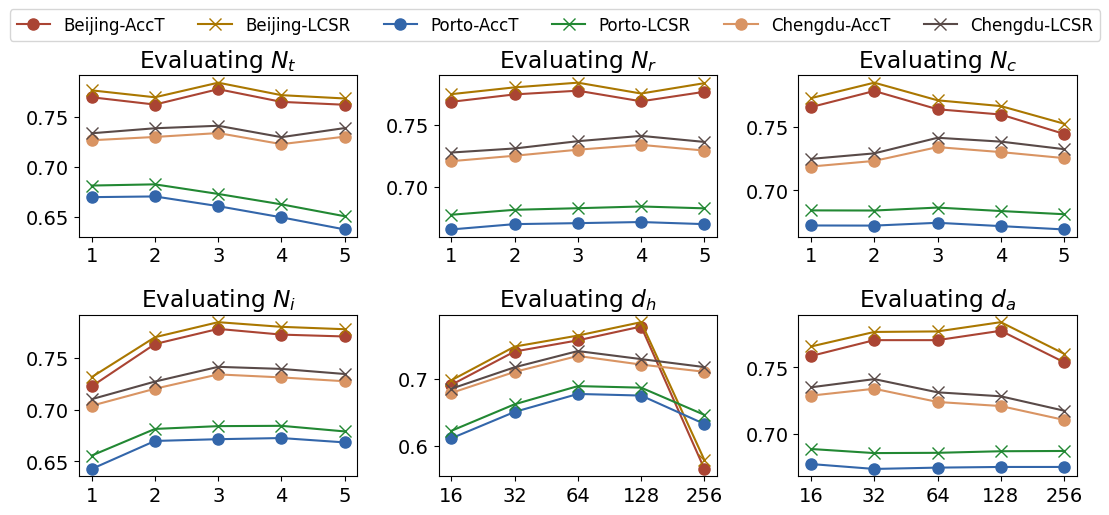}
  \vspace{-0.15in}
  \caption{AccT \& LCSR vs. Hyper-parameters}
  \vspace{-0.15in}
  \label{fig:hype}
\end{figure}
We consider the following hyper-parameters: (1) the number of recurrent neural network (RNN) layers ($N_t$ and $N_r$) in trajectory encoding module and road encoding module; (2) the number of graph convolution network (GCN) layers $(N_c)$ and graph isomorphism network (GIN) layers $(N_i)$; (3) the settable-dimension size of RNN hidden representations for trajectory and road $(d_h)$; (4) the settable-dimension size of attention mechanism $(d_a)$. In particular, given a hyper-parameter, we first select its value range according to the experience under some constraints (e.g., the limitation of GPU memory). Then, we conduct experiments on the validation set of Beijing, Porto and Chengdu to determine its optimal value. By default, we conduct experiments at a sampling rate of 50\%. As shown in Fig.~\ref{fig:hype}, we plot the AccT and LCSR for different hyper-parameters. In summary, we set each hyper-parameter with the value corresponding to the optimal performance as follows: (1) For Beijing, we have $N_t = 3$, $N_r = 3$, $N_c = 2$, $N_i = 3$, $d_h = 128$, $d_a = 128$. (2) For Porto, we have $N_t = 2$, $N_r = 4$, $N_c = 3$, $N_i = 4$, $d_h = 64$, $d_a = 16$. (3) For Chengdu, we have $N_t = 3$, $N_r = 4$, $N_c = 3$, $N_i = 3$, $d_h = 64$, $d_a = 32$.

\subsection{Effectiveness of Loss Weight and Partition}
\label{sec:6.3}
\begin{figure}
  \centering  
  \includegraphics[width=0.7\linewidth]{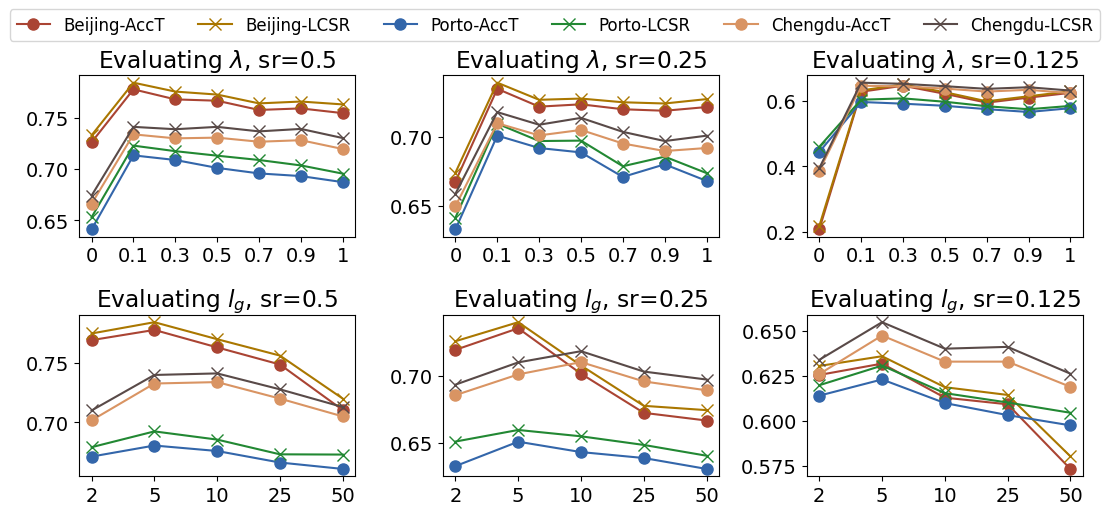}
  \vspace{-0.15in}
  \caption{AccT \& LCSR vs. the loss weight $\lambda$ and the grids side length $l_g$}
  \vspace{-0.15in}
  \label{fig:loss&length}
\end{figure}

To fine-tune the loss weight $\lambda$, we vary it from 0 to 1 when training. We compute the two metrics for the validation data. The result is shown in Fig.~\ref{fig:loss&length}, from which we can observe a noticeable improvement in model performance as $\lambda$ varies from 0 to 0.1. The improvement in performance proofs the effectiveness of our alignment module, particularly beneficial for scenarios with extremely sparse trajectory points. Based on the majority voting rule, the best values of $\lambda$ for Beijing, Porto and Chengdu are both 0.1, which is set as the default values in subsequent experiments.

Additionally, we further optimize the grid partition by adjusting the side length $l_g$ of grids, the outcomes are illustrated in Fig.~\ref{fig:loss&length}. In essence, the size of grids represents the granularity of regions, which determines the precision and generalization of the trajectories. Specifically, finer granularity can provide more precise GPS coordinates, while coarser granularity can enhance the generalizability of the trajectory. By default, the grids side length of Beijing and Porto are optimally set to 5m, and for Chengdu it is set to 10m.

\begin{table*}[ht]
\centering
\caption{Effectiveness Results on Test Data}
\vspace{-0.1in}
\label{tab:performance}
\setlength{\tabcolsep}{2.5pt}
\scalebox{0.59}{
\begin{tabular}{|c|cc|cc|cc|cc|cc|cc|cc|cc|cc|}
  \hline
  \multirow{3}{*}{} & \multicolumn{6}{c|}{\textbf{Beijing}} & \multicolumn{6}{c|}{\textbf{Porto}} & \multicolumn{6}{c|}{\textbf{Chengdu}}\\
  \cline{2-19}
  {} & \multicolumn{2}{c|}{\textbf{50\% (30s)}} & \multicolumn{2}{c|}{\textbf{25\% (60s)}} & \multicolumn{2}{c|}{\textbf{12.5\% (120s)}} & \multicolumn{2}{c|}{\textbf{50\% (30s)}} & \multicolumn{2}{c|}{\textbf{25\% (60s)}} & \multicolumn{2}{c|}{\textbf{12.5\% (120s)}} & \multicolumn{2}{c|}{\textbf{50\% (30s)}} & \multicolumn{2}{c|}{\textbf{25\% (60s)}} & \multicolumn{2}{c|}{\textbf{12.5\% (120s)}}\\
  \cline{2-19}
  {} & \textbf{AccT} & \textbf{LCSR} & \textbf{AccT} & \textbf{LCSR} & \textbf{AccT} & \textbf{LCSR} & \textbf{AccT} & \textbf{LCSR} & \textbf{AccT} & \textbf{LCSR} & \textbf{AccT} & \textbf{LCSR} & \textbf{AccT} & \textbf{LCSR} & \textbf{AccT} & \textbf{LCSR} & \textbf{AccT} & \textbf{LCSR} \\
  \hline
  \textbf{MDP} & 50.18\% & 0.5172 & 49.19\% & 0.5081 & 46.93\% & 0.4730 & 52.07\% & 0.5281 & 48.69\% & 0.5003 & 47.35\% & 0.4831 & 51.05\% & 0.5174 & 46.80\% & 0.4788 & 45.21\% & 0.4611 \\
  \hline
  \textbf{HMM} & 53.55\% & 0.5462 & 51.43\% & 0.5290 & 51.19\% & 0.5226 & 52.58\% & 0.5359 & 50.51\% & 0.5209 & 50.25\% & 0.5077 & 52.93\% & 0.5388 & 50.10\% & 0.5125 & 46.61\% & 0.4780 \\
  \hline
  \textbf{FMM} & 54.38\% & 0.5471 & 53.29\% & 0.5431 & 51.37\% & 0.5281 & 53.59\% & 0.5391 & 52.13\% & 0.5290 & 48.91\% & 0.4964 & 53.12\% & 0.5390 & 52.40\% & 0.5319 & 48.95\% & 0.5003 \\
  \hline
  \textbf{AMM} & 58.31\% & 0.5917 & 57.21\% & 0.5819 & 51.85\% & 0.5289 & 56.22\% & 0.5693 & 55.83\% & 0.5649 & 51.03\% & 0.5227 & 57.72\% & 0.5843 & 56.13\% & 0.5700 & 50.41\% & 0.5122 \\
  \hline
  \textbf{MTrajRec} & 61.54\% & 0.6352 & 55.92\% & 0.5797 & 47.06\% & 0.4971 & 57.53\% & 0.5912 & 55.77\% & 0.5641 & 52.27\% & 0.5399 & 60.57\% & 0.6172 & 57.48\% & 0.5881 & 50.89\% & 0.5207  \\
  \hline
  \textbf{L2MM} & 62.39\% & 0.6355 & 57.98\% & 0.5951 & 52.40\% & 0.5367 & 56.41\% & 0.5759 & 55.51\% & 0.5649 & 53.76\% & 0.5488 & 59.19\% & 0.6021 & 56.90\% & 0.5801 & 50.10\% & 0.5149  \\
  \hline
  \textbf{GraphMM} & 66.01\% & 0.6768 & 61.96\% & 0.6302 & 55.06\% & 0.5616 & 59.68\% & 0.6172 & 57.07\% & 0.5860 & 52.33\% & 0.5406 & 62.14\% & 0.6328 & 60.31\% & 0.6142 & 53.44\% & 0.5455 \\
  \hline
  \hline
  \textbf{NC} & 51.37\% & 0.5247 & 48.29\% & 0.4955 & 41.76\% & 0.4291 & 44.67\% & 0.4576 & 41.50\% & 0.4266 & 38.71\% & 0.3993 & 53.29\% & 0.5413 & 50.61\% & 0.5174 & 42.53\% & 0.4341 \\
  \hline
  \textbf{NI} & 44.31\% & 0.4521 & 43.27\% & 0.4429 & 38.82\% & 0.3956 & 33.21\% & 0.3489 & 32.39\% & 0.3345 & 30.89\% & 0.3191 & 45.56\% & 0.4667 & 43.91\% & 0.4503 & 41.54\% & 0.4237 \\
  \hline
  \textbf{NM} & 69.66\% & 0.7036 & 68.72\% & 0.6956 & 60.48\% & 0.6135 & 59.24\% & 0.6007 & 57.47\% & 0.5841 & 54.93\% & 0.5603 & 63.28\% & 0.6409 & 60.13\% & 0.6129 & 52.98\% & 0.5388 \\
  \hline
  \textbf{\modelName} & \textbf{77.78\%} & \textbf{0.7843} & \textbf{73.53\%} & \textbf{0.7398} & \textbf{63.19\%} & \textbf{0.6360} & \textbf{68.05\%} & \textbf{0.6924} & \textbf{65.12\%} & \textbf{0.6599} & \textbf{62.31\%} & \textbf{0.6308} & \textbf{73.39\%} & \textbf{0.7412} & \textbf{71.02\%} & \textbf{0.7183} & \textbf{64.75\%} & \textbf{0.6569} \\
  \hline
  
\end{tabular}
}
\vspace{-0.1in}
\end{table*}

\subsection{Effectiveness Comparison}
Apart from comparing \textbf{\modelName} with baseline methods, we replace our \textbf{\modelName} by three variations to conduct ablation study, namely \textbf{NC}, \textbf{NI}, and \textbf{NM}, to evaluate the effectiveness of different parts of encoding in \textbf{\modelName}. In \textbf{NC}, we use fully connected (FC) layers to replace graph convolutional network in trajectory encoding module. In \textbf{NI}, we use FC layers to replace graph isomorphism network in road encoding module. In \textbf{NM}, we replace the road-to-trajectory mapping mechanism with raw graph node coordinates. Note that the effectiveness of our proposed trajectory-road representation alignment module is evaluated in Sec.~\ref{sec:6.3}.

Table~\ref{tab:performance} reports the results of all methods with different sampling rate, and we have the following observations:

\noindent (1) Traditional rule-based offline methods (i.e., \textbf{MDP}, \textbf{HMM}, \textbf{FMM}) perform the worst because they are designed to model offline complete trajectories and are unable to learn general patterns from sparse and large-scale online scenarios, lacking robustness.

\noindent (2) When examining the results of \textbf{NC}, \textbf{NI}, \textbf{NM}, and \textbf{\modelName}, we observe that the graph neural networks in the encoding part are crucial, which validates the effectiveness of our graph structure in addressing the trajectory-road heterogeneity. Furthermore, using road representations as the initial representations for nodes in the trajectory transition graph to bridge the representation interaction also further enhances performance, as \textbf{NM} performs worse.

\noindent (3) \textbf{\modelName}~exhibits the best performance across all metrics. For instance, \textbf{\modelName}~outperforms the best existing methods (e.g., \textbf{GraphMM}) by 17\% on AccT for the test data of Beijing at a sampling rate of 50\%. In Porto, the best baseline method, GraphMM, achieves lower accuracy at the highest sampling rate than our method does at the lowest sampling rate. This demonstrates that, by using our method, service providers can reduce the sampling rate of trajectories to save costs while still achieving better performance than existing methods, as our method maintains strong performance even at lower sampling rates.

\noindent (4) When comparing metrics across different datasets, the methods perform similarly on Beijing and Chengdu, with slightly better results than that on Porto. We attribute this primarily to the Porto dataset’s higher road density, which results in a greater number of road segments close to the correct one, making the matching task more challenging.

\noindent (5) The LCSR is consistently better than the AccT across all methods. By definition of LCSR and AccT, AccT penalizes every incorrect match equally while LCSR emphasizes the accuracy of the entire sequence match, so when there are matching errors, the decline in AccT will be more pronounced than in LCSR. Moreover, this indicates that most map matching methods are not prone to frequent alternations between correct and incorrect matches, which possess good global matching capabilities and fault tolerance.

\noindent (6) As the sampling rate decreases, the performance of all methods declines, with the impact being more significant for learning-based methods due to the inevitable influence of data sparsity on the quality of the models' learning. 


\subsection{Efficiency Comparison}
\setlength{\tabcolsep}{3pt}
\begin{table}
\centering
\caption{Efficiency of Different Methods}
\vspace{-0.1in}
\label{tab:efficiency}
\scalebox{0.65}{%
\begin{tabular}{|c|ccc|ccc|ccc|}
  \hline
  \multirow{2}{*}{} & \multicolumn{3}{c|}{\textbf{memory size (MB)}} & \multicolumn{3}{c|}{\textbf{training time (s)}} & \multicolumn{3}{c|}{\textbf{matching time (s)}}\\
  {} & \textbf{Beijing} & \textbf{Porto} & \textbf{Chengdu} & \textbf{Beijing} & \textbf{Porto} & \textbf{Chengdu} & \textbf{Beijing} & \textbf{Porto} & \textbf{Chengdu}\\
  \hline
  \textbf{MDP} & 1819MB & 2039MB & 2122MB & - & - & - & 389.14s & 361.15s & 599.51s  \\
  \textbf{HMM} & 1209MB & 1388MB & 1361MB & - & - & - & 427.97s & 380.05s & 589.08s \\
  \textbf{FMM} & 897MB & 931MB & 981MB & - & - & - & 1.13s & 1.02s & 1.87s \\
  \textbf{AMM} & 957MB & 1013MB & 1124MB & - & - & - & 3.42s & 3.05s & 5.16s \\
  \textbf{MTrajRec} & 9045MB & 12428MB & 11265MB & 182.4s & 2200.2s & 25672.4s & 51.22s & 42.27s & 73.68s\\
  \textbf{L2MM} & 9087MB & 11875MB & 10898MB & 189.1s & 2314.2s & 27032.2s & 6.71s & 5.26s & 9.10s\\
  \textbf{GraphMM} & 8537MB & 11752MB & 10378MB & 48.4s & 645.2s & 7311.4s & 8.06s & 6.96s & 11.18s\\
  \textbf{\modelName} & 2530MB & 2299MB & 2357MB & 11.9s & 126.4s & 1507.8s & 1.09s & 0.95s & 1.65s\\
  \hline
\end{tabular}}
\vspace{-0.15in}
\end{table}

For efficiency evaluation, we record the memory usage, training time and inference time at a sampling rate of 50\%. The memory usage represents the required memory size for applying corresponding methods, which is used to evaluate the memory efficiency. The training time is used to evaluate the offline learning efficiency for neural network based methods. In particular, we compute the average time of an epoch for each method. \textcolor{black}{The matching time (i.e., inference time) can evaluate the online matching efficiency.} Specifically, we record the time for each method to match 10K trajectories at a sampling rate of 0.5. Note that for rule-based methods running on the CPU, we apply parallel mechanisms to ensure fairness compared to learning-based methods running on the GPU with batch.
The outcomes of these evaluations are presented in Table~\ref{tab:efficiency}. From our observations, we deduce the following:

\noindent (1) Rule-based methods (i.e., \textbf{MDP}, \textbf{HMM}, \textbf{FMM}, and \textbf{AMM}) require less memory than learning-based methods (i.e., \textbf{MTrajRec}, \textbf{L2MM}, \textbf{GraphMM} and \textbf{\modelName}) because they do not need to keep network structures in memory. Among learning-based methods, \textbf{\modelName} is the most memory efficient.

\noindent (2) Offline map matching methods, such as \textbf{MDP}, \textbf{HMM}, \textbf{MTrajRec}, \textbf{L2MM} and \textbf{GraphMM}, which have not been specifically optimized for efficiency, perform poorly in terms of online matching efficiency. This demonstrates that the naive strategies for adapting offline methods to online scenarios are unsuitable. Additionally, among all HMM-based models, the efficiency of \textbf{HMM} itself is significantly lower than other models due to its lack of improvements to enhance efficiency. Furthermore, the lack of support for online scenarios in naive \textbf{MDP} and \textbf{HMM} models leads to repeated invocations, making their matching time in online scenarios almost unacceptable for practical applications.

\noindent (3) Among all learning-based methods, our model exhibits the best offline learning efficiency. Furthermore, among all methods, our model demonstrates the best online matching efficiency. Compared to the method that specifically designed for online scenarios (i.e., \textbf{AMM}), our model achieves more than $3\times$ improvement in online matching efficiency, \textcolor{black}{making it highly practical and feasible for real-world applications.} 
This is attributed to the seamless integration of our designed online MDP and reinforcement learning methods, which effectively extract key information while dynamically updating and utilizing historical data. Our approach thus avoids redundant computations to achieve high efficiency.

\noindent (4) When comparing results across different datasets, training one epoch takes significantly longer on the Chengdu due to the higher number of trajectories. For online matching time, Chengdu requires more time as well, primarily because its trajectories are generally longer than those in Porto and Beijing.

\begin{figure}
  \centering  
  \includegraphics[width=0.83\linewidth]{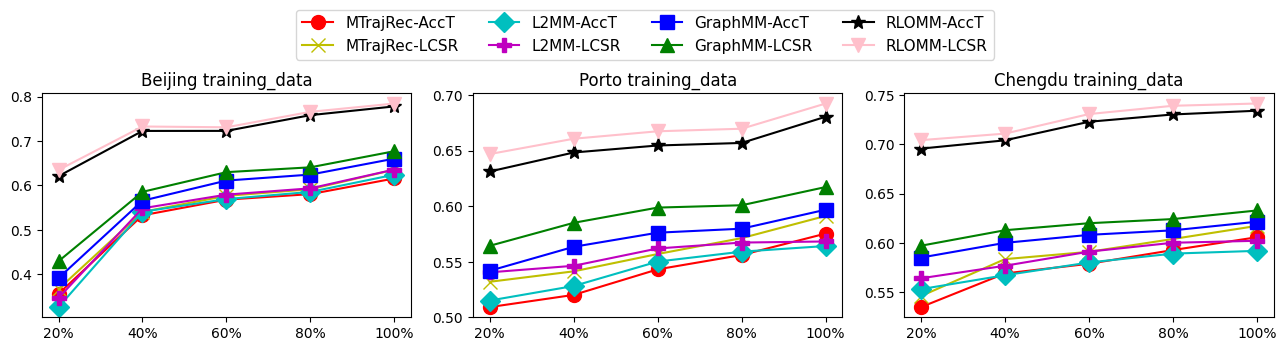}
  \vspace{-0.15in}
  \caption{AccT \& LCSR vs. the Scalability.}
  \vspace{-0.15in}
  \label{fig:scalability}
\end{figure}
\subsection{Scalability Comparison}
To compare the scalability of the learning based methods (i.e., \textbf{MTrajRec}, \textbf{L2MM}, \textbf{GraphMM}, and \textbf{\modelName}), we train different models by varying the training data size. In particular, we sample 20\%, 40\%, 60\%, 80\% and 100\% from the training data, and collect the associated AccT and LCSR of the online prediction over the test data at a sampling rate of 50\%. From Fig.~\ref{fig:scalability}, we have the following observations:

\noindent(1) All methods perform better with an increased amount of training data, since a larger dataset encompasses a wider range of situations, allowing the model to learn more effectively.

\noindent(2) Our \textbf{\modelName} consistently achieves the best performance. Notably, in many instances (e.g., the AccT and LCSR metrics for Beijing), as the sampling rate decreases, the performance gap between our \textbf{\modelName} and other methods widens.

\noindent (3) The performance degradation varies across datasets due to differences in the amount of trajectory data. Beijing, with its relatively smaller number of trajectories, experiences the most significant decline, whereas Chengdu shows notably less degradation.

\subsection{Case Study}
\begin{figure}
  \centering  
  \includegraphics[width=0.83\linewidth]{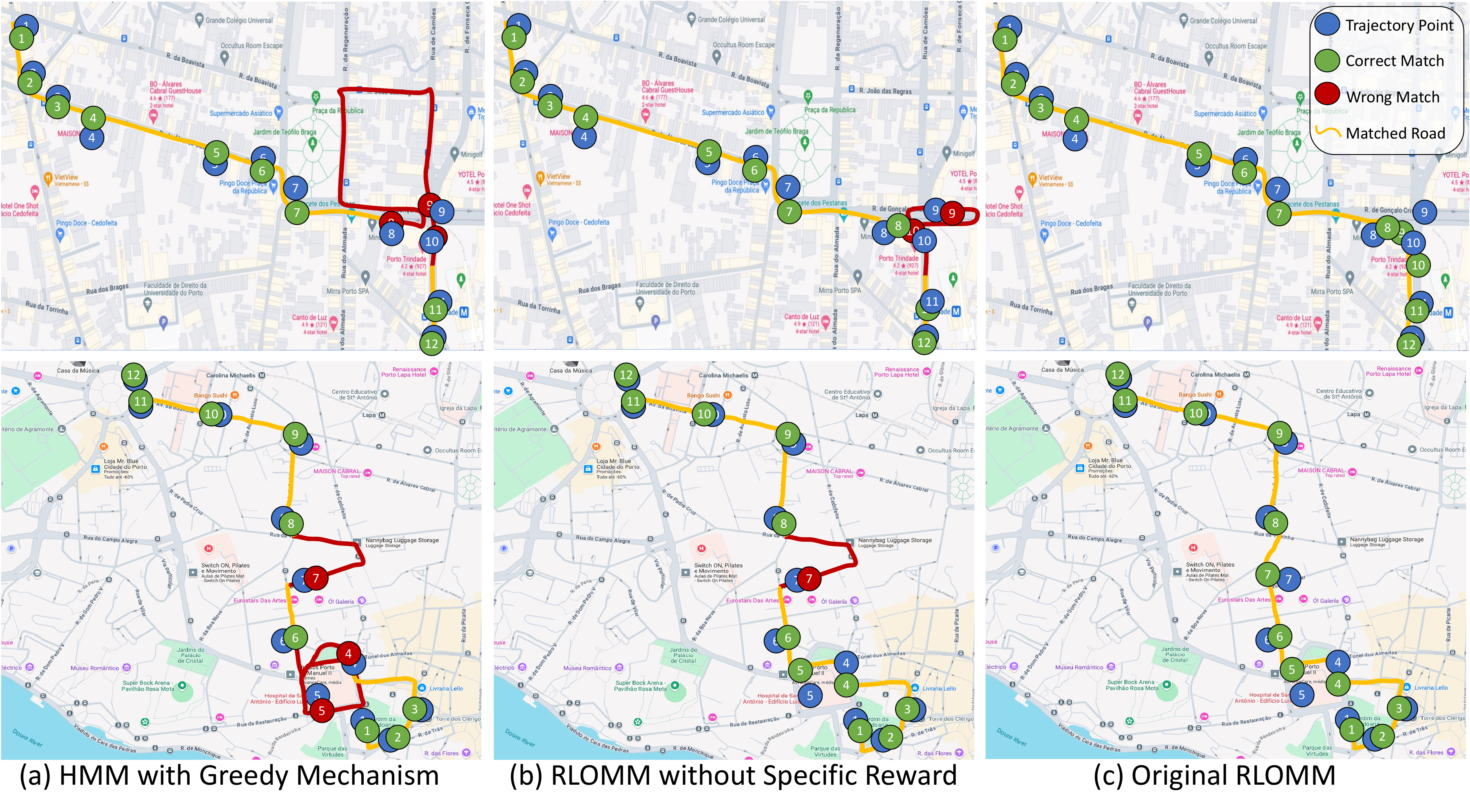}
  \vspace{-0.15in}
  \caption{An illustration of matching result for two trajectories of Porto. The blue points are the trajectory points need to be matched, the green points are the correct matches and the red points are the wrong matches.}
  \vspace{-0.15in}
  \label{fig:case}
\end{figure}

To visually illustrate the matching result and further demonstrate the superiority of our method over the greedy approach, we conduct a case study. Specifically, we select two trajectories from Porto to compare the matching performance of three methods: the \textbf{HMM} with greedy mechanism, \textbf{RLOMM} without detour penalty and road connectivity reward designed in Sec.~\ref{sec:5.2}, and the original \textbf{RLOMM}. As shown in Fig.~\ref{fig:case}, we have the following observations:

\noindent (1) For HMM with a greedy mechanism, its matching results often exhibit incorrect detours because of selecting closer road segments to the trajectory point (e.g., the 9th point of the first trajectory, and the 4th and 5th point of the second trajectory). In particular, such greedy-induced matching errors often impact subsequent matches. For example, to maintain road continuity, the incorrect match of the 4th point in the second trajectory directly leads to the incorrect match of the 5th point. This inherent flaw in greedy methods results the unsuitability of the HMM approach for online scenarios.

\noindent (2) Comparing RLOMM with and without the detour penalty and road connectivity reward reveals that, thanks to the carefully designed rewards, our reinforcement learning-based framework can consider each match from a global perspective. For instance, in the case of the first trajectory, since the erroneous matching of the 9th trajectory point is not connected to the preceding 8th matched road, the road connectivity reward guides the model to avoid this incorrect matching. Furthermore, when matching the 10th trajectory point, the detour penalty encourages the model to consider previously unmatched road segments, leading to a correct match. The various rewards we design work in conjunction to guide the model in providing accurate matching results when faced with complex and error-prone scenarios.

\section{Related Work}
Map matching problem can be classified to two broad categories~\cite{mmsurvey}: \textit{Offline Map Matching} and \textit{Online Map Matching}. For \textit{Offline Map Matching}, researchers focus on matching complete trajectories in offline scenarios, which aims for optimal matching route with less constraint on processing time~\cite{survey_magzine}. For \textit{Online Map Matching}, trajectory points are continuously sampled and processed in a streaming manner, which imposes high demands on real-time performance.
\subsection{Offline Map Matching}
Early research on the offline map matching problem was primarily based on the concept of similarity comparison, that is, by defining similarity to identify the road segments most similar to the target trajectory. The authors in~\cite{currentmm, icde12, trajSeg} design algorithms with different measure of similarity (e.g., spatial distance, longest common subsequence) to find the most similar road segment as the matching result. 

When maps and trajectories are complex, similarity-based methods often perform poorly. To enhance performance, Hidden Markov Models (HMM) and HMM-enhanced models are used. HMMs assume that the current state depends on the previous one and consider the correct road segments as hidden states of the trajectory points, which are the observations.
Newson et al.~\cite{HMM} are the first to use HMM to address the offline map matching problem. 
Yang et al.~\cite{FMM} integrate HMM with precomputation to achieve high processing speed. 
The authors in~\cite{MDPMM} use MDP to address the limitations of HMM, designing dynamic outlier resolution and improved preprocessing functions for map matching.
Moreover, considering both the spatial structures and the temporal constraints of the trajectories, \cite{stmatching, impstmatching, ifmatching} design methods that characteristic relationships between GPS points.
However, although HMM is widely used in commercial software, its performance in low-sampling-rate scenarios is poor, limiting its potential for cost reduction and efficiency improvement~\cite{praGuide}.

Recently, deep-learning-based map matching methods have attracted the attention of researchers. 
In order to learn general patterns from trajectory data, \cite{deepmmzhao, deepmmfeng} build deep learning based models to utilize all the trajectory data for joint training and knowledge sharing. 
The authors in~\cite{mtrajrec} propose MTrajRec to recover the fine-grained points in trajectories and map match them on the road network in an end-to-end manner. 
Jiang et al.~\cite{l2mm} propose L2MM to generate high-quality representations of low-quality trajectories through high-frequency trajectory enhancement. 
Liu et al.~\cite{graphmm} present a graph-based approach that incorporates conditional models to leverage various road and trajectory correlations. 
As for cellular trajectory map matching, \cite{LHMM, FL-AMM, DMM2} design different learning methods (e.g., multi-relational graph learning, federated learning, heuristic reinforcement learning optimizer) to learn the mapping function from cellular trajectory points to road segments. As remote sensing imagery is innovatively applied to various spatiotemporal data tasks~\cite{POIjn,urbancmx}, it is conceivable that the map matching task could similarly leverage diverse types of novel data in the future.

\subsection{Online Map Matching}
For online map matching, research primarily focuses on the Hidden Markov Model (HMM) or enhanced versions of HMM, focusing on improving efficiency.

Considering noise and sparseness of GPS data, Goh et al.~\cite{ohmm} propose an enhanced HMM with a Variable Sliding Window method to integrate spatial, temporal, and topological information to map the GPS trajectories to the road network in real time. 
To further address complex and changeable city traffic conditions, Liang et al.~\cite{onlinelearning} design OLMM, which uses online learning to improve accuracy without requiring any prior human labeling.
\cite{fu2021online} proposes an online map matching algorithm based on a second-order HMM with an extended Viterbi algorithm and a self-adaptive sliding window mechanism to enhance trajectory data processing accuracy in complex urban road networks.
To cope with high noise, Jagadeesh et al.~\cite{2017TITSonline} design a HMM to generate partial map-matched paths in an online manner, and use a route choice model to reassess each HMM-generated partial path along with a set of feasible alternative paths.
Furthermore, the authors in~\cite{routepre} propose an algorithm that replaces future GPS points with a probabilistic route prediction model.
Recently, Hu et al.~\cite{AMM} design an adaptive online map matching algorithm called AMM, which calibrate GPS observation data for various measurement errors and under complex urban conditions. 
Additionally, in addressing the issue of cellular trajectory map matching, which involves handling trajectories based on cellular positioning data, \cite{AccReal} design an incremental HMM algorithm that combines digital map hints and a number of heuristics to reduce the noise and provide real-time estimations. 

However, aforementioned online methods are limited by their rule-based nature, which prevents them from fully capturing the spatio-temporal correlations~\cite{hmmAna}. Additionally, these methods are predominantly greedy in their matching process, ignoring the impact of current matches on future outcomes, leading to poor robustness and accuracy~\cite{mmsurvey, surveyVMMT}.

\section{Conclusion}
This paper introduces \textbf{\modelName}, a reinforcement learning framework designed for the online map matching problem, offering high efficiency and robustness tailored for real-world applications. We have addressed three key limitations present in existing methods: low efficiency, poor robustness, and insufficient handling of trajectory-road heterogeneity. Initially, we modeled the problem as an Online Markov Decision Process with efficient integration of real-time and historical information. Next, we adopted reinforcement learning and meticulously designed rewards to enhance dynamic adaptability and robustness. Additionally, we incorporated an effective graph structure for trajectories and roads to address the heterogeneity, facilitating better fusion between them. Moreover, we proposed a trajectory-road representation alignment module to enhance the robustness of representations and reduce their distance in latent space. Extensive evaluations on three real-world datasets have confirmed the efficacy of \textbf{\modelName}.

\begin{acks}
This work was supported by the National Natural Science Foundation of China (No.62425203, No.62032003, No.U21B2016). Haitao Yuan is the corresponding author of the work.
\end{acks}


\bibliographystyle{ACM-Reference-Format}
\bibliography{ref}

\end{document}